\documentclass[10pt,amssymb,twocolumn,notitlepage]{article}

\usepackage{epigraph}
\usepackage[affil-it]{authblk}
\usepackage{dcolumn}
\usepackage[stable]{footmisc}
\usepackage{blkarray}
\usepackage{bm}
\usepackage[mathscr]{euscript}
\usepackage[font=scriptsize]{caption}
\usepackage{subcaption,graphicx}
\usepackage[table]{xcolor}
\usepackage[margin=1.7cm]{geometry}
\usepackage{tcolorbox}
\usepackage{hyperref}
\usepackage{gensymb}
\usepackage[font=footnotesize]{caption}
\hypersetup{
    colorlinks=true,
    linkcolor=blue,
    filecolor=magenta,      
    urlcolor=blue,
}
\usepackage{steinmetz}
\usepackage{amssymb}
\usepackage{makecell}
\usepackage{algorithm,lipsum,xcolor,caption}
\usepackage{enumitem}
\usepackage{amsmath,esint}
\usepackage{fancyvrb}
\usepackage{xcolor}
\usepackage{arydshln}
\usepackage{mathtools}

\definecolor{newblue}{rgb}{0.0, 0.28, 0.67}
\definecolor{newgreen}{rgb}{0.13, 0.55, 0.13}
\definecolor{newred}{rgb}{0.87, 0.72, 0.53}

\definecolor{newblue}{rgb}{0.0, 0.28, 0.67}
\definecolor{newgreen}{rgb}{0.13, 0.55, 0.13}
\definecolor{newred}{rgb}{0.87, 0.72, 0.53}
\newcommand{\R}{\mathbb{R}}

\title{An Ample Approach to Data and Modeling }
\author{Luciano da Fontoura Costa \\ \emph{luciano@ifsc.usp.br}}
\affil{S\~ao Carlos Institute of Physics -- DFCM/USP} 
\date{20th Sept 2021, v.1 
\\          12th Oct 2021, v.2
}

\begin{document}

\twocolumn[
\begin{@twocolumnfalse}
    \maketitle
    \begin{abstract}
    Developments in Science and technology have relied extensively on modeling and pattern
    recognition.  In the present work, we describe a framework for modeling
    how models can be built that integrates concepts and methods from a wide range of fields.
    The information schism between the information in the real-world and that
    which can be gathered and considered by any individual information processing agent
    is characterized and discussed, which is followed by the presentation of 
    a series of the adopted requisites while developing the reported modeling approach. The issue 
    of mapping from datasets into models is subsequently addressed, as well as some of the
    main respectively implied difficulties and limitations.  Based on these considerations, an
    approach to meta modeling how models are built is then progressively developed.
    The reference $M^*$ meta model framework is presented first, which relies
    critically in associating whole datasets and respective models in
    terms of a strict bijective association.  Among the interesting features of this model
    are its ability to bridge the gap between data and modeling, as well as paving
    the way to a paired algebra of both data and models which can be employed
    to combine models in hierarchical manner.  After illustrating the
    $M^*$ model in terms of patterns derived from regular lattices, the reported modeling
    approach continues by discussing how sampling issues, error and overlooked data
    can be addressed, leading to the $M^{<\epsilon>}$ variant.  The frequent and important
    situation in which the data needs to be represented in terms of respective probability
    densities is treated next, yielding the $M^{<\sigma>}$ meta model, which is then
    illustrated respectively to a real-world dataset (iris flowers data).   Several 
    considerations about how the developed framework can provide insights about 
    data clustering, complexity, causality, network science, collaborative research, deep learning, 
    and creativity are then presented, followed by overall conclusions.
     \end{abstract}
\end{@twocolumnfalse} \bigskip
]

\setlength{\epigraphwidth}{.49\textwidth}
\epigraph{`\emph{So much closer on the lake, the new star.}'}
{\emph{LdaFC.}}


\section{Introduction}

Despite the many intricacies that characterize our world and the universe, 
there are some properties that seem to underlie in a more shared, fundamental, and 
systematic manner the structure and dynamics of natural phenomena.  Among them, 
we have the two following intriguing facts:  (i) every portion of our world is strongly 
interconnected and interdependent one another along time and space; and (ii) at the 
same time, there  are \emph{severe} limitations to the information that can be collected
and processed by any computing system, be it natural (e.g.~humans) or artificial (e.g.~digital 
computers).  

Perhaps necessary and unavoidably, these two important principles directly 
oppose one another, establishing an interesting mutual global/local duality or tension
which may have well been, as will be suggested in the present work, at the very 
core of the appearance  and unfolding of life and intelligence themselves.   On one
side we have a vastness of global interconnections extending through an impressive
range of scales, on the other the extremely limited resources of every kind available
to any known information processing system, including living beings.
This intriguing duality is here deemed here  to important and inexorable enough as to warrant the
name of \emph{information schism} which, though related to the semantic gap,
implies a wider context.

The level of interdependency of natural structures and phenomena is so breathtaking
as not to be often realized.  The following two examples should suffice to
materialize this issue.   First, consider the dynamics of a pendulum on a table
in front of us.  Because of the gravitational force always established between any
two bodies of mass, and as a consequence of this force fields extending to infinity,
the movement of this pendulum will be affected by  \emph{every single bit of mass in the universe}.  
Though this influence will certainly become smaller and smaller as the distance between the
masses increases, it will nevertheless be present and therefore influence, even as
an infinitesimal element in the background noise, in limiting full
accuracy to be achieved for the measurements and predictions about the pendulum dynamics.  
Our second example, which is related to interactions along time, consists of the 
relatively smaller and localized events that reverberate through lasting periods of time, as it was possibly the case of the meteor that led to dinosaurs being extinct, therefore allowing mammals
to evolve. 

Interestingly enough, both the above issues become particularly critical in non-linear systems
--- which encompass virtually a wide range  of real-world phenomena --- because these can amplify 
even minute perturbations into major effects.  Other critical limitations of information 
processing by agents include, but are not limited to, the fact that several objects and states 
are non-observable in nature, the limited available \emph{time} for any action, as well as 
the possible presence of error and noise in every measurement taken from the real world.  

So it is that nature has somehow self-organized itself into a never ending web of 
interconnections and interdependences extending along every possible scales of
space and time.  From the perspective of nature itself, this does not constitute a problem,
because the physical world does not seem to require help of external computing resources.

However, with the appearance of individual agents, such as living beings, the information 
schism acquired much more relevance, representing a major challenge to be circumvented 
in some manner.  The main problem here is  that agents --- be it natural or artificial, individual 
or collective --- can only exchange and process minute portions of mass, energy and 
information.  As the survival and eventual reproduction of these agents rely critically
on their interaction with the environment (which may also include other agents), so as to make 
suitable decisions with basis on predictions, it becomes mandatory
that any of these agents incorporates the adequate means for receiving, processing, predicting, 
and acting on the environment.  

\begin{figure}[h!]  
\begin{center}
   \includegraphics[width=0.5\linewidth]{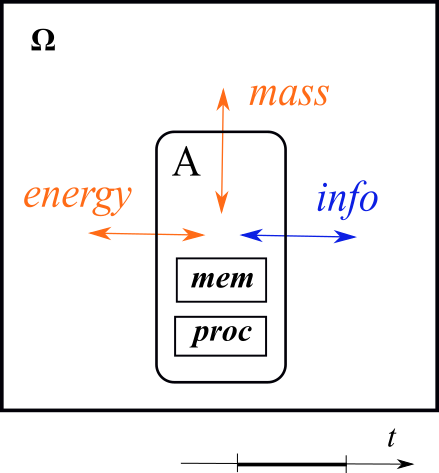}  
    \caption{An individual agent $A$ in an environment $E$ exchanges and acts upon
    mass, energy and information.   The survival and reproduction of this agent depends critically
    on taking timely decisions by processing the information received from the
    environment and then acting suitably on it, while taking in account previous experiences
    stored in some manner.  Agents capable of information 
    processing can be modeled as incorporating memory and processing capabilities,
    which are both finite and require some physical expense in terms of mass and
    energy, being consequently limited.  There are also constraints regarding the period
    of time it takes for predicting and making decisions, and it is important to keep in mind
    that the environment $E$ almost certainly also contains other information 
    processing agents representing potential threat or collaboration.}
    \label{fig:individual}
    \end{center}
\end{figure}

When contemplated from the above discussed perspective, the continuing existence of 
living beings and other information processing entities can be realized as being truly 
phenomenal.  After some additional reflection, we may perceive that the success of individual 
living beings  has largely relied on taking timely decisions on what to do based on information sampled 
from the respective environment, while also taking into account previous experiences
consolidated through some kind of memory such as nervous systems or biochemical dynamics
characteristic of each species.  Figure~\ref{fig:individual} illustrates the basic condition of an
individual $A$ in an environment $E$.

A more in-depth view of the concept of decision making is particularly critical for our formulation.  
It should be recalled that the taking of effective decisions relies critically at least on the following 
aspects: (i) sampling enough information from the environment; (ii) taking into account 
previous experience, especially in the sense of possessing comprehensive understanding 
of the environment; and (iii) identifying the situation as corresponding to some instance
of already experienced problem; and (iv) having the means for making accurate
 \emph{predictions} guiding the possible decisions.  More informally, models could ben 
understood as magic mirrors reflecting not only the real-world properties, but also their
respective consequences.

An intriguing relationship can thus be established between the action of taking a decision 
and the basic principle underlying science, namely the \emph{construction of models} by 
using the scientific method (e.g.~\cite{CostaModeling,CostaCreat}).

The scientific method also relies importantly on \emph{obtaining quantified information} about 
the phenomenon of interest, as well as previous related knowledge, in order not only to 
\emph{better understand} that phenomenon, but also to \emph{make accurate predictions} 
about it.  Thus, in essence, decision making by information processing entities can be directly 
related to the development of models.   
Thus we posit that \emph{decision taking} by agents is basically 
the same as \emph{scientific modeling}, sharing not only the same objectives, but also 
the mechanisms for achieving results.    In a sense, a model can be understood as a 
logic-mathematic-computational construct involving conditions that need to be satisfied 
by the observed data.  

Interestingly, a further relationship can be established between individual 
decision taking with  the area of \emph{pattern recognition} 
(e.g.~\cite{Koutrombas, CostaPatt, DudaHart}).    In this area of great current interest,
the main objective is, given a set of properties or features measured from an object, to reach a conclusion
about its possible \emph{class} or category.  There are two main types of pattern recognition:
\emph{supervised} and \emph{non-supervised}, the former being characterized by availability of 
previous knowledge, examples or prototypes of the existing categories, which are not
available in non-supervised classification.  In both cases, and especially 
in the  supervised case, a direct parallel can be established between decision taking/ model building 
with pattern recognition.  The obtention of experimental measurements in modeling can
be directly associated with the derivation of properties of the entities to be classified, the
consideration of previous experience/models is reflected in the information available about the
classes, and the obtained prediction can be directly paired with the action of making predictions.

The intrinsic relationship between decision making, pattern recognition and modeling can 
also be inferred from the evolutionary perspective that endowed humans (as well as other 
living beings) with these two abilities.  The point here is that, in case pattern recognition and 
modeling are distinct, they would require different neuronal and cognitive respective abilities,
which is much more expensive than sharing the same neuronal resources for addressing
both these critically important tasks.

We could go much further because several other actions such as urban planning, land
management, education, economic and social policies, to name but a few, may also be 
related with modeling, decision taking, or pattern recognition.  Remarkably, also the field of 
arts can be related to the modeling framework by understanding an art piece with a dataset 
and the model with the conditions estimated to be necessary for positive respective appreciation 
and/or impact.   Ultimately, it becomes difficult to find a human activity that can not be somehow
related to model building and/or pattern recognition.  
  
The above discussed relationship between decision taking, modeling, and pattern recognition
probably represents the most important and critical characteristic of the approach discussed in
the current work.  For at least the following reason: this allows us to incorporate concepts and 
methods from a wide range of related scientific fields, especially philosophy of science, artificial intelligence (e.g.~\cite{Russell_Norvig}) 
and pattern recognition, discrete mathematics, statistics, physics, 
complex networks, and data science, to name but a few.

Another important feature of the reported approach regards its logic-mathematic-computational
formalization of the activity of model building through a \emph{meta-model}, which can 
allow us to better understand and draw more objective and general results and conclusions 
concerning the properties, advantages and limitations of model building.

Though new models can be obtained in a never ending
number of ways, here we focus on a methodological framework based on logical combinations of the
existing models while considering set operations between the respective datasets.  More
specifically, instead of logically combining models (which can also be done) irrespectively of data, the
considered method also allows expressing the dataset of interest as a combination of other
existing datasets, then obtaining the sough model as a logical combination of the models that
can be associated to the respectively identified datasets.  It is shown that, by establishing
an bijective association between the datasets of interest and respective models it becomes
possible to obtain a bridge between these two domains, with each set operations being used to
manipulate datasets becoming bijectively associated with a respective logical operation.

Given the data and model realms, it is possible to start with some data of particular importance
and then look for a model, or vice-versa.  An interesting asymmetry seems to characterize
the development of science through model constructing, residing in the fact that though
new models can be obtained by logic, exact combination of the existing models, these models
would still need to be associated to some real world data, which corresponds to its respective
physical validation.  Interestingly, the proposed bijective association between datasets and
models avoids this situation, because every combined model will necessarily be
associated to a respective dataset.

In spite of its idealizations in several respects, this first meta-model, which is henceforth referred to as 
the $M^*$ model, provides a sound reference for better understanding more 
realistic modeling through the progressive incorporation of characteristics such as
noise, incomplete sampling and/or characterization, classification errors, etc.  

In this work, the basic overall of the $M^*$ framework is also extended to address problems
like sampling, error, noise (yielding the $M^{<\epsilon>}$ meta model), and to take into
account the stochastic representation of the data in terms of respective probability densities,
therefore leading to the $M^{<\sigma>}$ meta model.  Both the $M^*$ and $M^{<\sigma>}$
approaches are illustrated by respective case-examples.

To complement work, we discuss how the concepts and methods related to the developed
modeling frameworks can provide insights about areas of great current importance including
clustering, complexity, collaborative research, deep learning, and creativity.

\section{Specifying the Problem}  \label{sec:requisites}

It often happens that the difficulties in developing a solution to a given problem ultimately derive from 
lack or imprecisions while specifying  the respectively sought goals and constraints.  
Thus, it is reasonable to initiate the development of the approach reported in this work by listing 
the many requirements and characteristics that were initially specified.

The main objectives and constraints that have been adopted in the currently described approach
are listed in the following:
\vspace{0.3cm}

\noindent \textbf{[R1]}  - Allow the integration of several related concepts such as modeling, decision making, pattern recognition, etc.;

\noindent \textbf{[R2]}   - Integrate the progressive incorporation of knowledge that characterizes scientific advance;

\noindent \textbf{[R3]} - Accommodate the tension between specificity in modeling datasets bijectively
and the generality implied by every data element in those sets non-injectively satisfying the same associated model;

\noindent \textbf{[R4]} - Allow the representation of data elements in terms of respective features
(measurements), as it is typical in pattern recognition;

\noindent \textbf{[R5]}  - Allow the identification of the main challenges in modeling and other
related areas;

\noindent \textbf{[R6]}  - Provide subsidies for better understanding clustering, 
complexity, complex networks,
ontologies, collaborative science, deep learning and creativity, among other possibilities;

\noindent \textbf{[R7]}  - Adhere to both model- and data-driven perspectives;

\noindent \textbf{[R8]}  - Lead to an effective modeling methodology that can be 
eventually automated  in software and/or hardware engines;

\noindent \textbf{[R9]} - Be relatively formal but remain nevertheless
accessible, while also maintaining good didactic potential;

\noindent \textbf{[R10]} - Allow the incorporation of stochasticity related to dataset and modeling;

\noindent \textbf{[R11]}  - Allow the incorporation of the tuning role of parameters in 
scientific modeling;

\noindent \textbf{[R12]}  - Pave the way to compositions of models, in the sense that the
modeling results can be feedbacked as input or into other modeling systems;

\noindent \textbf{[R13]}  - Be as congruent as possible with the human understanding of
modeling and pattern recognition, as well as many of the involved concepts;

\noindent \textbf{[R14]} - Allow multiple data elements to be queried simultaneously, as
motivated by modeling, and still provide good performance when applied to
single individuals.

\section{Mapping Datasets into Models}


As approached in this work, the basic operation in modeling is considered to be the mapping
of datasets into respective models.  As such, it is important to discuss this operation in more
detail, which consists the main objective of this section.  More specifically, we will develop
a reasoning allowing an \emph{bijective association} to be established between the datasets
and the respective models.   Recall that, mathematically, an bijective association consists of
an binary association of elements belonging to two sets that has the properties of being 
\emph{reflexive}, \emph{symmetric} and \emph{transitive}.  bijective associations are particularly
important because they can be understood as implementing a network of \emph{causal}
relationships between the several involved components.

For simplicity's sake and for all subsequent purposes in this work, this type of relationship
may be understood as establishing  a \emph{identity} or bridge between the dataset and
model domains.  

Other important issues include the understanding of how
\emph{parameters} can be accommodated into models, the need to quantify the properties of
the data elements into respective \emph{features or properties}, as well as the several 
possible \emph{types of models}, not to mention the several manners in which
models can be progressively developed.  Therefore, it is hoped that the concepts and discussions
developed in this section contribute a sound basis for building the sought meta model, 
as well as for identifying and discussing the possible limitations while mapping datasets into 
models, which will be addressed in the subsequent section.  By \emph{meta modeling}
it is henceforth understood the endeavor of modeling how models are built and developed.

\begin{figure}[h!]  
\begin{center}
   \includegraphics[width=0.8\linewidth]{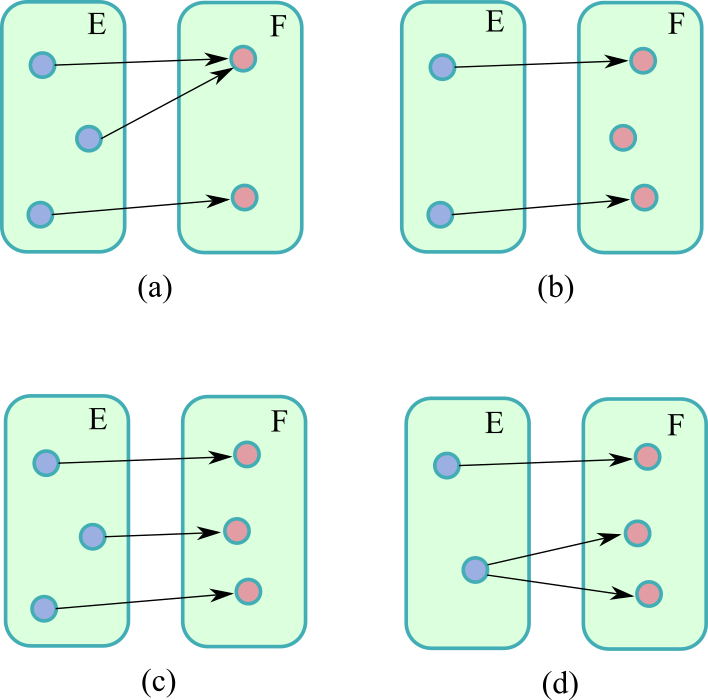}  
    \caption{Four possible types of mappings, here understood in the  mathematical
    context, from datasets in $E$ into models in $M$.  (a): \emph{non-injective} mapping, meaning
    that more than one dataset is associated to a same model.  (b) \emph{non-surjective} mapping,
    in which one model in $M$ is not verified for any dataset in $E$.  (c) The \emph{bijective} situation
    in which each dataset is associated to a single model, and all models end up associated datasets.
    The situation in (c) is critically important because it means that the mapping can be inverted.
    Formally speaking, though the mapping shown in (d) cannot be characterized as a function
    (though this restriction in relaxed in some approaches),  it does
    represent a relatively common situation while mapping data into models and can be easily
    addressed by merging those datasets that map into the same model, yielding an injective map. }
    \label{fig:sets}
    \end{center}
\end{figure}

We start by presenting, in Figure~\ref{fig:sets}, four types of mappings that
may take place between the datasets in the environment $E$ and the respective models in the
model framework $M$.

In Figure~\ref{fig:sets}(a) we have a \emph{non-injective} mapping, in which more than one dataset 
$\omega$ of $E$ are mapped into the same model $m$ in $M$.  Though it can be understood that both 
these datasets are explained by that model, it is impossible to distinguish between the two original
datasets from their respective image.    A \emph{non-surjective} mapping is illustrated in 
Figure~\ref{fig:sets}(b), in which some
of the models in $M$ have not been verified respectively to any of the existing datasets in $E$.  
This situation could be informally understood as a ``model in wait for a dataset''.  This situation is
also unwanted because we have models that are not verified.  The situation depicted
in Figure~\ref{fig:sets}(c) corresponds to a \emph{bijective} mapping between the elements
of  $E$ into $M$, being therefore \emph{invertible} unlike the two previous situations. 

The critical importance of adopting a bijective, invertible mapping of datasets into models resides
in the fact that this type of relationship both avoids the ambiguity of a non-injective mappings as
well as the existence of unverified models.    

A forth situation is worth consideration, and it has to do with mappings that, by not adhering to the
usual concept of mathematical function, allow more than one model to be associated to 
a same dataset,
as illustrated in In Figure~\ref{fig:sets}(d).  This situation can be addressed by understanding 
that those multiply satisfied models actually correspond to the same model, yielding a respective
injective map.   As we will see, this type of mapping is also relatively common regarding data 
elements, but cannot occur when an bijective association is to be established between datasets 
and models.  This situation may also be caused by  insufficient sampling of data or errors.

Interestingly, while in our approach the mapping between datasets and models is henceforth 
understood as an bijective association, all the \emph{data elements} inside each dataset $\omega_i$ map in a non-injective manner into the same model $m_i$, being therefore \emph{not}
subjected to an bijective association.   At the same time, any data element may belong
to more than one dataset.

The fact that greater freedom of mapping is allowed in the case of data elements is actually welcomed 
because it disentangles the seemingly opposite requirements in modeling and pattern recognition
respectively to having specificity of models regarding whole sets of data, 
but generalization of models with respect to individual data elements.
From the pattern recognition perspective, it means that all datasets in a given $\omega_i$ belong
to the same category defined by the respective model $m_i$,  which seems to be quite reasonable.

Scientific models can be understood as involving \emph{variables}, \emph{constants}, and 
\emph{parameters}, among other possible components.  
Variables include all quantities that may vary during an experiment;
constants refer to quantities that never vary during or between experiments; and parameters
are quantities that may vary from an experiment to another.  Variables are often subdivided
as being dependent and independent (or free).   As implied by its name, a variable is
said to be dependent in case it is expressed in terms of the others in a given model.
It is interesting to observe that the concept of variable dependence is relative to each
specific model, because a variable that is dependent in one case my be independent
in another model.

In the case of a simple pendulum, we have time as a free variable and  the angular position and 
speed as variables dependent of time.  The mass of the bob and the length of the rod correspond
to parameters.  The gravity acceleration can very probably be taken as a constant, given that
it is difficult to change its value in a laboratory.   

These three main types of modeling elements can be immediately associated with pattern recognition
concepts:  variables are the measurements (or features) of the data; constants are constants, 
while parameters correspond to 
adjustments influencing the measurements or decisions.  For instance, in a neuronal network
the parameters would correspond to the weights and bias of each neuron, or could refer to 
the smoothing level adopted while simplifying images.  In a physical model, the parameter
tuning allow a specific phenomenon to fit the respective model.

The proper setting of parameters is critical for modeling and pattern recognition, since they
directly influence the decisions.  In the present work, we understand that the parameters are
always adjusted so as to guarantee the bijective association between datasets and
respective models.  This adjustment can be made through some \emph{optimization} procedure, 
varying the parameter values so as to lead to no errors in the decision.  The more frequent 
situation of possible decision errors will be addressed in Section~\ref{sec:errors}.  It is also
possible to search for suitable parameters value during data analysis and model building.

So far, we have assumed that the data elements in each dataset can be directly operated by the
model in order to verify the respective adherence.  However, the practical analysis of a given dataset
by a model is, in general, impossible unless the elements in this dataset have been first properly
represented in terms of a set of categorical or quantitative \emph{features}, also called
measurements, characteristics, attributes, and properties, in the pattern recognition area.  Therefore,
a \emph{further level of mapping} needs to be incorporated into modeling and pattern recognition,
extending from datasets into feature sets.  

\begin{figure}[h!]  
\begin{center}
   \includegraphics[width=0.7\linewidth]{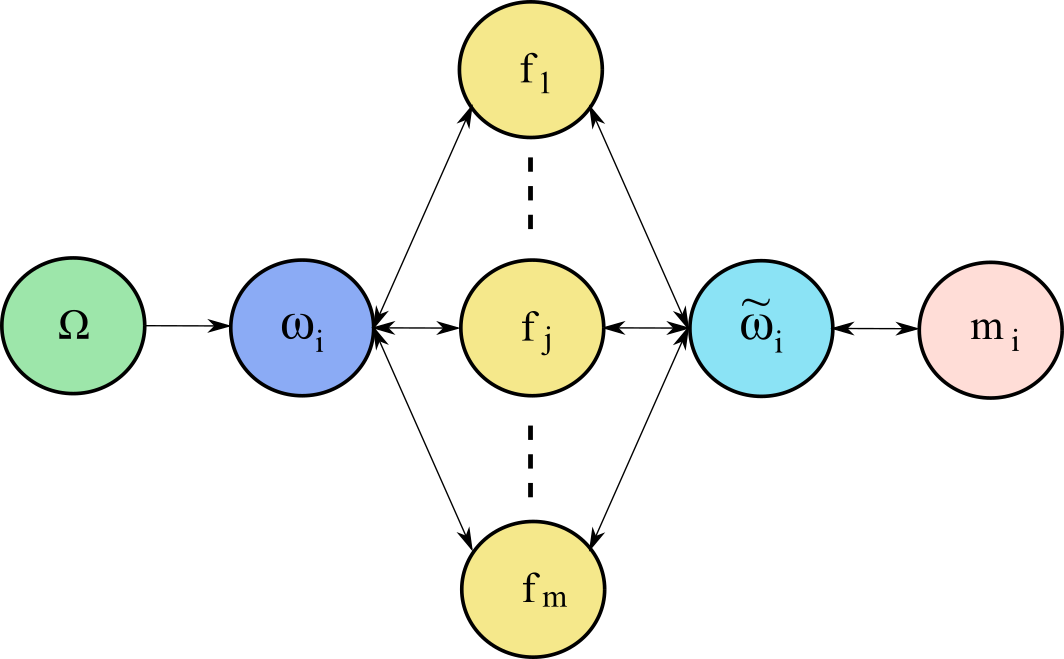}  
    \caption{The modeling or recognition of a dataset $\omega_i$ derived from a respective universe
    $\Omega$ requires each data element to be first mapped into a set of categorical and/or quantitative 
    features $f_j$, $j = 1, 2, \ldots, m$, which defines in a bijective manner a new dataset $\tilde{\omega}_i$.
    For simplicity's sake, the feature-associated datasets $\tilde{\omega}_i$ will not be shown in
    the other diagrams in this work.}
    \label{fig:features}
    \end{center}
\end{figure}

The diagram in Figure~\ref{fig:features} illustrates how a given dataset $\omega_i$ can be mapped into respective features $f_j$, $j = 1, 2, \ldots, m$, which are then 
sent to the respective model $m_i$.

Observe that the description of a dataset $\omega_i$ therefore give rise to a transformed
version of that dataset $\tilde{\omega}_i$, upon which the models can now objectively operate.  For 
simplicity's sake, the latter type of datasets will be omitted from other figures and diagrams
of this work, but they will be nevertheless understood to be present.

It is not often realized that features are \emph{always} present in typical decision, modeling
and recognition tasks, including the measurements representing the very entities of interest.
For instance, before we can decide that the presented entity is or not a dog, it needs to be
transformed into an image by our visual systems, or typically transformed into respective
matrices in an artificial system.  Features can be transformed and combined
in endless manners, but the results can always be understood as features.  

As we learn from the pattern recognition area, it often constitutes quite a challenge to select a proper
set of features describing the analyzed entities.  Observe that the number of features involved
in a model defines a respective multidimensional \emph{feature space}.  Ideally, each data 
element should be mapped into a common feature space in a bijective manner, so as to 
establish an bijective association between the data elements and their representations in 
terms of the considered features.  It is also particularly important to identify the \emph{smallest}
set of features that may allow a problem to be reasonably solved.

Though the features are assumed to provide a complete, invertible representation of the
datasets in our first approach ($ M^*$), which is necessary to maintain the bijective
association between data elements and respective models, it is also possible to subsequently
adapt this same model for situations 
when the features no longer provide an invertible mapping with the data
elements.  For consistency of modeling, we also assume that all data elements in $E$
are always characterized in terms of every considered and applicable features.  Observe that
the fact of a data element not allowing the derivation of a feature considered in a given
existing model $m_i$ may automatically eliminate the possibility of that element satisfying that
model, in case the dataset cannot be described in terms of other features. 
At the same time, a feature could be missed that represents the only manner to
discriminate between two distinct datasets.

We have so far addressed several points related to the data elements, datasets, types
of mapping of the latter into models, and features. Now, we approach the modeling
level itself.

It is important to keep in mind that any model can be 
immediately associated with a
decision or categorization, namely that of the dataset satisfying or not the model, or
to which an extent it adheres to the dataset.
There are several types of possible models/decisions:  thresholds, rules, equations, descriptions, etc.
Any of these may be involved in the henceforth considered modeling.

It is also interesting to divide the possible models into two major groups: (a) those
that seek an \emph{optimal} (minimum or maximum of some merit figure); and those
aimed at achieving a given property within a reasonable margin of accuracy.  While
the former type of problems is directly related to the ample and important area of
mathematical optimization, the latter involves defining some margin of tolerance
and working with probabilities. 

Merit or fitness figures can be associated to each obtained model, reflecting the requirements
specific to each problem.   Possible merit figures include the length of the model description,
its intelligibility to humans, and the cost of checking if a dataset satisfies a model, among
many other possibilities.  A particularly interesting objective is, given a new dataset, to find
the largest dataset entirely containing it, as this would account for the most general explanation of that
dataset.  In this case the larger dataset will nevertheless have to be restricted if one wants to
keep the bijective association.  Also, unless the smaller dataset has some special significance, 
it could be therefore subsumed into the larger model.  Several such simplifications and specifications 
are allowed by the proposed meta modeling approach.  Another particularly interesting situation 
concerns, given a set of datasets to find interrelationships between them.

\section{Limitations in Mapping Datasets into Models} \label{sec:errors}

Several components of our meta-modeling approach have been presented and discussed
in the previous section.  Now, we address some of the most common types of limitations and
constraints related to those components.

Given a dataset $\omega_i$ of interest, it is possible one or more its data elements to have been
assigned by mistake, or that other data elements be missing.  In these cases we will have
a dataset that does not fully correspond to our expectations.  Let's illustrate this situation in terms
of the following example.  Let $\omega$ be a dataset that has been singled out for modeling
as a consequence of having its data elements associated with a posited new plant species.  
Spurious samples from other species may be included in $\omega$, while other samples of the
considered species are overlooked, e.g.~by some sampling procedure.  
These situations will imply in 
inconsistencies leading to incorrect model being identified for that dataset, and
probably lead to less accurate and incomplete model identification and combination.

Missing data elements are characteristic of the sampling that is unavoidably required
in case of infinite or too large sets of data elements.  

Errors may also occur while mapping datasets into features.  These may include mistakes, finite
resolution or noise while measuring the features.  It is also possible that the equations or
program used to estimate the features is intrinsically incorrect, leading to improper characterization.
Errors taking place while measuring or calculating features can severely impact the identification
of a valid model for the given dataset.   Another related problem concerns the fact that a
feature that is critically necessary for obtaining a model for a given dataset is overlooked or unknown.

Another possible source of errors takes place at the modeling level itself.  Here, we may have
inconsistent decisions defined in terms of the features, logical errors,
or the overlooking of some important features. 

An important type of error not often realized in modeling and pattern recognition is the situation
in which some of the data elements in the data environment $E$ have not yet
been checked  respectively to every existing model, which may also undermine the obtained 
results.

\section{The M* Meta-Model} \label{sec:M_star}

Having discussed some of the main aspects and components involved in
mapping from datasets into models,
as well as possible respective limitations, we are now in position of developing a
more principled and relatively formal meta-model that can account for as many of the
requirements listed in Section~\ref{sec:requisites} as possible.

\begin{figure}[h!]  
\begin{center}
   \includegraphics[width=0.8\linewidth]{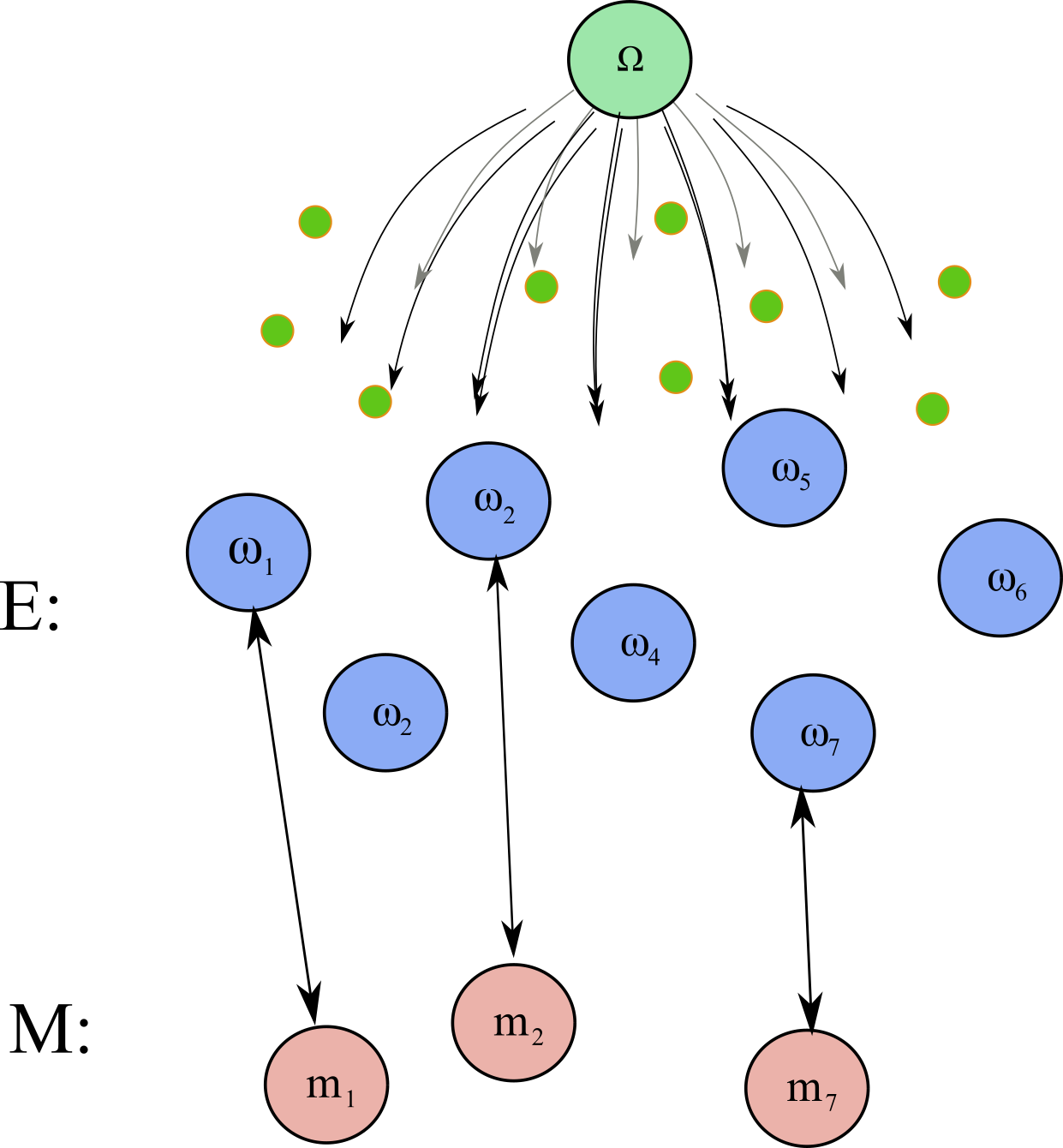}  
    \caption{The  $M^*$ meta model, decision taking, and pattern recognition.  
    The universe set $\Omega$ contains all possible data
    elements (small green circles) $x_j$, which can be successively drawn into environmental datasets 
    $\omega_i$, $i = 1, 2, \ldots, N_{\omega}$, so that $\omega_i \subset \Omega$, which
    define the current data environment $E$.    Each of these datasets 
    may eventually become associated to a respective model $m_i$ explaining necessarily every possible 
    element of $\omega_i$.    The set of existing models is understood to correspond to the
    modeling framework $M$.  The adherence between the data elements in the available datasets 
    and the existing models needs to be continuously updated in order to ensure overall consistency.
    The critically important bijective association between datasets and models precludes one
    dataset of being associated to more than a model, and vice versa, but this case can
    also be easily accommodated into the $M^*$ framework.}
    \label{fig:pre_model}
    \end{center}
\end{figure}

We start with the overall structure depicted in Figure~\ref{fig:pre_model}, involving a finite 
number of possible data elements represented as a universe set $\Omega$.  
These basic data elements $x_i$ are henceforth assumed 
to be finite, with $j = -1, 2, \ldots, N_{\Omega}$.  Observe that the largest possible environment
$E$ corresponds to the \emph{power set} of $\Omega$, containing $2^(N_{\Omega})$ subsets.

The observable, or available, or restricted
set of subsets of $\Omega$ are understood to constitute the environment $E$ upon
which models can be built,
being therefore accessible as a set of datasets  $\omega_i$, $i = 1, 2, \ldots, N_{\omega}$, 
each of which are therefore composed by data elements.  The initial configuration of a 
given model framework can be associated to the respectively assumed \emph{postulates} 
or \emph{hypothesis}.

The data elements in $\Omega$ can be equiprobable (exist in the same number) or
not.  Interestingly, both cases can be identically addressed by the proposed framework,
though the non-equiprobable case will imply in some data elements to be less likely
(taking a long time) to be incorporated into $E$.    Probabilistic situations can be
approached by using the $M^{<\sigma>}$ framework to be described in a subsequent
section.

Observe that a same data element  may appear in more than one dataset, as it individually
may satisfy more than one model.  This property is reasonable and compatible
with our concept of modeling and recognition, because a same entity can indeed
satisfies several models.  For instance, a cat is a mammal, but also a mammal, and it has a tail. 
Each of these decisions are 
normally taken as valid categories, though we may be particularly interested in some more
specific or general property.

\begin{figure}[h!]  
\begin{center}
   \includegraphics[width=0.8\linewidth]{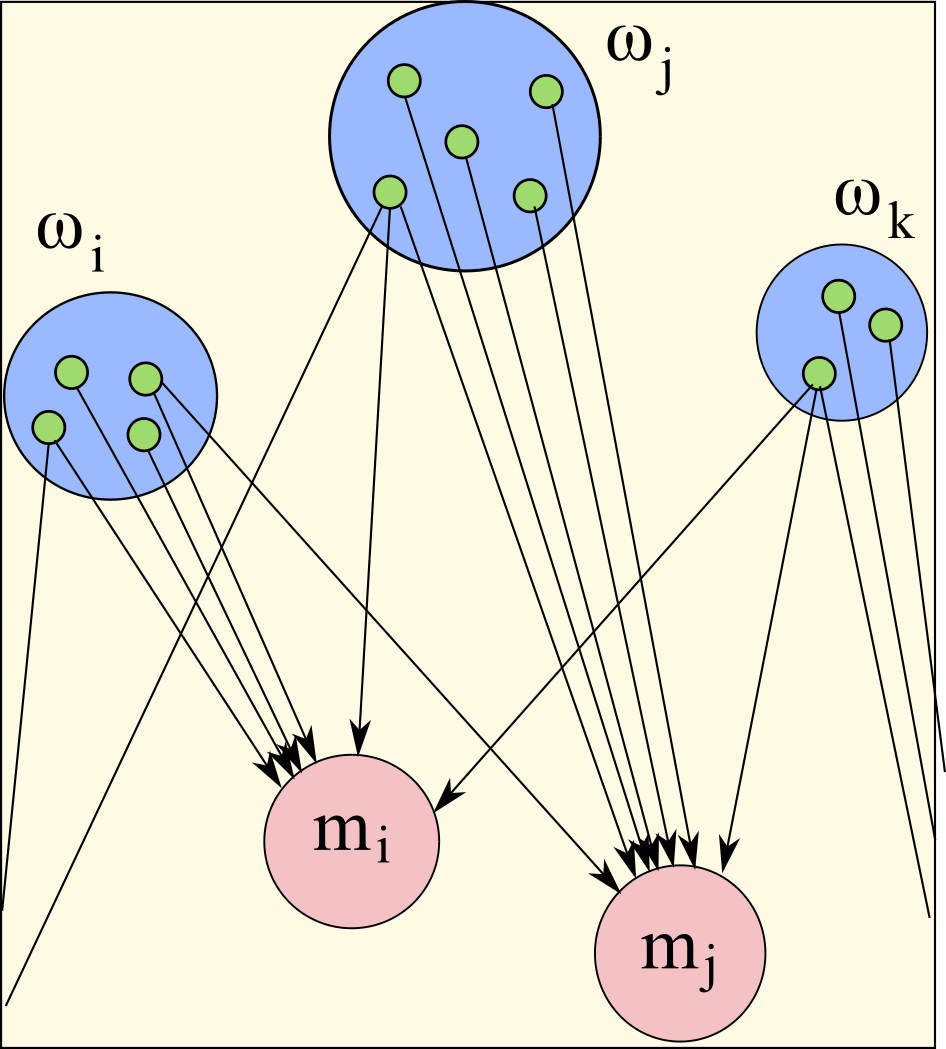}  
    \caption{Zooming into an hypothetical situation illustrates the fact that all data elements (green)
    in a dataset $\omega_i$ necessarily map into the same model  $m_i$ in a non-injective
    manner. Each of the arrows extending from a data element $x_j$ to a respective model
    $M_i$ indicates that $x_j$ satisfies $m_i$.   At the same time, a bijective mapped is ensured between each dataset and
    its respective model.  This feature of the proposed framework is essential for disentangling
    these two seemingly conflicting characteristics of modeling.  Though each of the data elements
    in a pair $(\omega_i,m_i)$ does satisfy the model $m_i$, this model is only understood
    to be  fully satisfied with respect to all the elements in $\omega$.
    The figure also shows that a same
    data element belonging to one of the existing datasets can map into more 
    than one model.  The features layer has been omitted for simplicity's sake.}
    \label{fig:zoom}
    \end{center}
\end{figure}

Figure~\ref{fig:zoom} presents a zoom of a hypothetical modeling situation, illustrating
some of the important features regarding the association between the dataset elements
and the respective models.

At each instant, the suggested meta model is understood to incorporate $N_M$ models $m_i$,
$i = 1, 2, \ldots, N_M$.  Each of the datasets $\omega$ may become associated to one and only 
one  respective model 
$m$, being henceforth understood that \emph{every} element of $\omega$ will satisfy the respective 
model $m$, and vice-versa, so that an \emph{bijective association} is consequently
established between each dataset and the respective model.    Observe that this aspect of
the $M^*$ framework actually defines two scales or levels of modeling, one at the data
element level, and another of higher hierarchy at the dataset level.

The current set of available datasets $\omega_i$ is henceforth called \emph{data environment} $E$,
while the existing models are henceforth understood to constitute the \emph{modeling framework} $M$.
The set containing all elements in any of the datasets $\omega_i$ of $E$ is henceforth
represented as $S_E$.

Taken jointly, these two sets may be related to the concept of \emph{current knowledge}.  Recall
that at any time, new data elements can be drawn from $\Omega$ and define new
datasets that can eventually assume enough importance in order to become subject of respective
modeling.  Examples of this possibility include but are by no means limited to the appearance
of a new species of living beings, the discovery of new stars, the birth of new individuals of a given
species, and the invention of new technological devices.  As addressed in more detail
in Section~\ref{sec:algebra}, it is also possible that two or more
existing datasets (or pairs dataset-model) be combined through set operations such as union,
intersection, complementation or difference.

In order to ensure consistence of proposed framework, the datasets are updated continuously,
in the sense that any new element drawn from $\Omega$ is checked respectively to each existing model
and incorporated into the associated dataset in case it satisfies that model before eventual
combination of models can be contemplated.
In addition, all the existing data elements are continuously 
checked respectively to any new incorporated model.

\begin{figure}[h!]  
\begin{center}
   \includegraphics[width=0.7\linewidth]{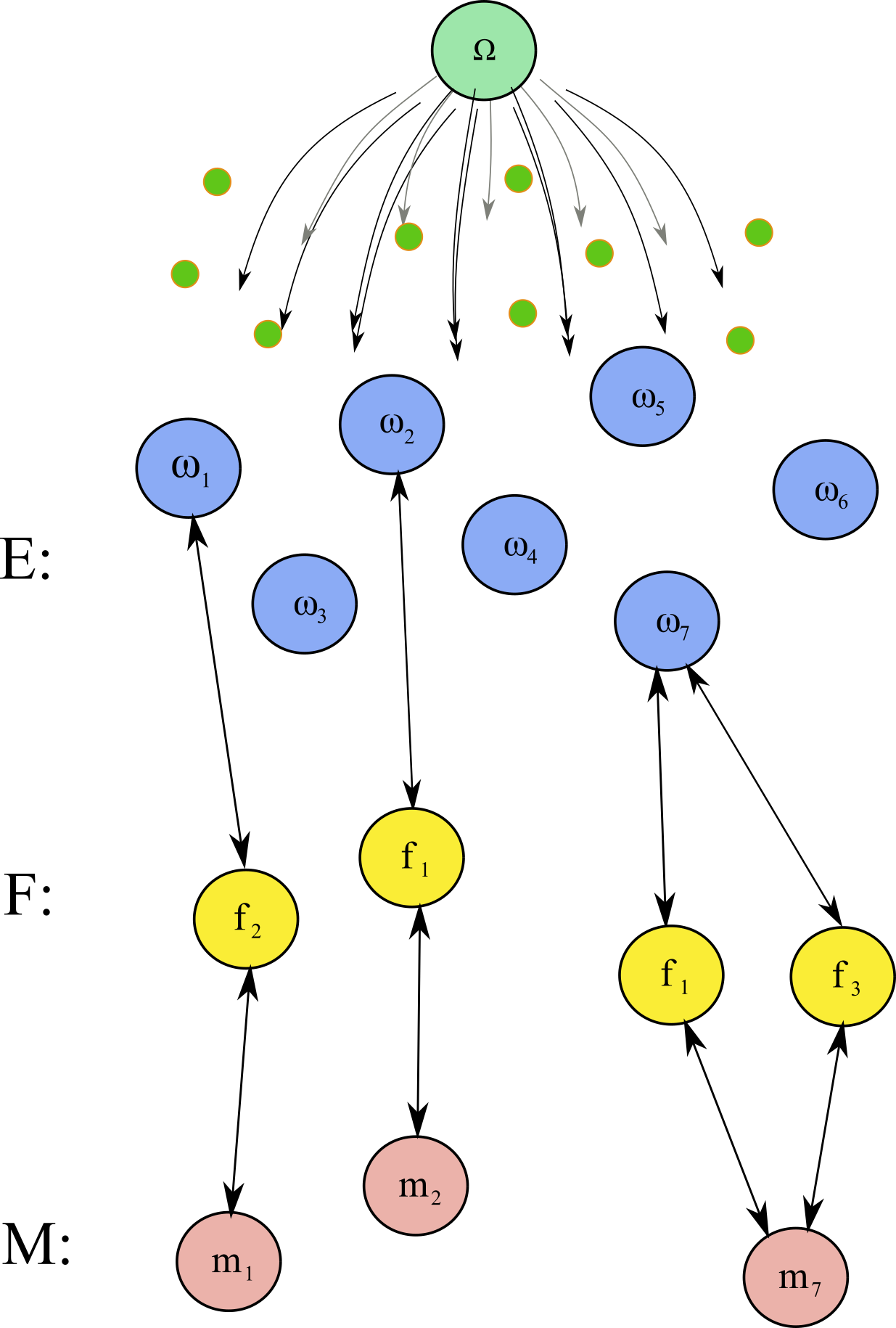}  
    \caption{The meta model $M^*$ expanded to incorporate the features respectively
    describing each of the datasets.  It should be recalled that models can only operate
    on data that has been categoric or quantitatively specified in an objective manner.
    In addition, more than one dataset be characterized in terms of a same type of
    feature.}
    \label{fig:model}
    \end{center}
\end{figure}

A more complete, expanded representation of the meta model $M^*$ is depicted in 
Figure~\ref{fig:model}, also incorporating the \emph{features} associated to each
dataset.

Observe that the bijective association between datasets and respectively associated models
is maintained by the new feature layer $F$, being reflected in the confluence of each dataset 
into a single model even
if multiple features were involved in the respective representation.  In addition, observe
that a same type of feature may be adopted for the characterization of more than one
dataset.    The incorporated layer consisting of features is henceforth understood to
constitute a new layer called the \emph{feature layer} $F$.

Given a \emph{bijective pairing} $(\omega_i,m_i)$, a decision can be immediately assigned
to the model, corresponding to its respective verification by the dataset.  Therefore, each
model corresponds to a question or decision, a feature that is directly related to the 
concept of causality.

There are other
interesting situations that can also be taken into account while integrating data and models
into the current framework.  For instance, when a new dataset $\omega_i$ is found to be entirely
contained in one of the existing datasets $\omega_j$, therefore being a subset of the latter, 
a \emph{restriction} of $\omega_j$ can be assigned to $\omega_i$.  For instance, let's say
that $\omega_i = \left\{ 1, 2, 3, 4, 5, 6, 7\right\}$ and $\omega_i = \left\{2, 3, 4\right\}$, 
the new dataset can be paired with a respectively restricted model corresponding to the
set operation $\omega_i = \omega_j -(\omega_j-  \omega_i)$.  Interestingly, this
sequence of reasoning implies \emph{recursion}, which can be approached by initially assigning
a provisory dummy model $m_i$.  

In fact, the association of a model to a dataset can be done as soon as the dataset is
given, through a labeling procedure that simply assigns an identifier (without semantic
content) to that dataset.  Evidently, this type of modeling does not contribute to the
modeling framework and cannot lead to a decision being respectively attached to the model
other than providing an identified for that dataset.

The above discussion suggests that assigning models to 
datasets that are subsets of the existing datasets may not only compound the modeling
framework, but also contribute to making it to increase in a combinatorial manner.
These cases can be more effectively addressed simply by understanding that a new
dataset is actually a subset of a larger dataset explained by a more comprehensive model.

Although the basic principle in the $M^*$ is to try to assign a model to every new dataset of
interest, cases in which the latter is already fully contained within the existing datasets
associated to models would only be justified in case the new datasets has some special relevance
requiring its discrimination, through a restriction, from the existing datasets.
Otherwise, this type of new dataset can simply be ignored.

The fact that a same data element can map into two or more models motivates the
use of measurements to quantify these interrelationships.  A possibility  consists simply
in considering the number $n_{i,j}$ of associations between each data element $x_{i,j}$ 
in a subset $\omega$ and all the existing models.  Then, relative frequency 
histograms (or respective moments) can be used to characterize each dataset (or
model).  Datasets leading to relatively large average number
of multiple connections $\left< n_{i,j} \right>$ can be understood as being less
\emph{specific}, in the sense that the respective data elements are strongly related,
though not through bijective relationships, with several models.   It is also possible to
count the number of individual data element connections received by a given model,
as this value provides insights about the generality of the models.  

It is equally interesting to study the distribution of these histograms among the several 
models in the  environment $E$.  In case the datasets are found to have similar
distributions of the r $n_{i,j}$ statistics, the modeling framework can be understood
as being more \emph{uniform}.   

Observe that the relationship between the datasets and respective models can also be represented
in terms of a \emph{bipartite} network having weights corresponding to the number of respective
data elements mapping from a dataset to a model.  Such networks can provide valuable information
about the overall structure, completeness, malleability, and robustness of the respective modeling 
framework.  It would be of particular interest to devise means for enhancing a given scientific
framework while taking into account the topological features of these networks.  These networks
should also present hierarchical structure reflecting not only the data/model hierarchy, but also
how the data elements are distributed amongst the existing models.

An analogue approach can be adopted regarding the types of features in a given $E$.

While the $M^*$ approach requires the bijective pairing of datasets and models, it is 
also possible to consider the following situation (and related variations).  Given a current
dataset environment $E$ and a modeling framework $M$, it may happen that one of the
datasets $\omega_i$ already paired with a respective model $m_i$, defining the pairing
$(\omega_i, m_i)$,  becomes associated
to another distinct model $m_j$.  This corresponds to the mapping situation depicted in
Figure~\ref{fig:sets}(d).  This is a case of particular interest because, though the mapping
from $\omega_i$ to $m_i$ is not typically considered a function, the inverse is a function.
These situations can be easily accommodated into the $M^*$ approach simply by
merging the two models $m_i$ and $m_j$ through their union or intersection, since
($\omega_i \cup \omega_i, m_i \lor m_j)$ and ($\omega_i \cap \omega_i, m_i \land m_j)$.

Reaching the most complete knowledge about $\Omega$ can be understood as the ultimate goal of
modeling.  This situation corresponds to having  models associated to every possible subset
$\omega \subset \Omega$.  Quite interestingly, this can be done in several manners, including
the following extreme approach: the model corresponding to each possible subset of 
$\Omega$ consists simply to enumerating its elements.  The problem with this trivial solution is
that not much is learned about the data elements and their grouping into datasets.  
Other approaches include the already discussed combination of existing models, as well
as developing completely new models based on insights provided by the similarity between
datasets by using the index $\Lambda$ suggested in Section~\ref{sec:errors}.

\section{A Paired Algebra of Datasets \emph{and} Models} \label{sec:algebra}

Though the $M^*$ meta model has so far been contemplated in a mostly static manner, additional
mechanisms may be incorporated allowing the progressive derivation of new models and datasets.  
A possible respective approach is described in the current section involving either combination of datasets
in terms of set operations or the integration of models by using logical connectives, which we
henceforth understood as a \emph{paired algebra} of datasets and models.  Yet another possibility
to be discussed elsewhere is the presentation of new pairs of datasets and models.

It should be also taken into account that the proposed framework
can be readily adapted to other types of models, e.g. by using production rules or composition
of functions as in neuronal networks.
Interestingly, the heuristics usually employed by humans for taking decisions seems to be
largely dependent of logical manipulations.  This property of the $M^*$ framework is related
to the fact that the consistency between models in always guaranteed in terms of the
respective data consistency.

Incidentally, the $M^*$ framework and its derivations
seems to be largely congruent with the way humans develop models, take decision and
perform pattern recognition.

The consistent combination of either datasets or models is immediately allowed by the
fact that the pairing between datasets and models corresponds to an bijective association,
which establishes a sound bridge  between these two important domains.
Under these circumstances, it immediately follows that set operations between datasets
$\omega$ become intrinsically linked to logical manipulations of respective models
$m$.  Some examples of the bijective associations between dataset
and modeling operations include:

\begin{align*}
   \textbf{Dataset Domain } E  &\Longleftrightarrow  \textbf{Model Domain } M \nonumber \\
   \omega_k = \omega_i    &\Longleftrightarrow  m_k =  m_i  \nonumber \\
   \omega_k   =  [\omega_i]^C    &\Longleftrightarrow  m_k = \neg m_i  \nonumber \\
   \omega_k   = \omega_i \cup \omega_j  &\Longleftrightarrow m_k = m_i  \lor m_j  \nonumber \\
   \omega_k   = \omega_i \cap \omega_j  &\Longleftrightarrow m_k = m_i  \land m_j  \nonumber \\
   \omega_k   = \omega_i - \omega_j  =  \omega_i \cap [\omega_j]^C   &\Longleftrightarrow m_k = m_i  \land \neg m_j  \nonumber \\
   \omega_k   = \omega_j - \omega_i  = [\omega_i]^C \cap \omega_j   &\Longleftrightarrow   m_k = \neg m_i  \land  m_j  \nonumber \\
   \omega_k   = [\omega_i \cup \omega_j]^C  = [\omega_i]^C \cap [\omega_j]^C  &\Longleftrightarrow   m_k = \neg m_i  \land \neg m_j  \nonumber \\
   \omega_k   = [\omega_i \cap \omega_j]^C  = [\omega_i]^C \cup [\omega_j]^C   &\Longleftrightarrow   m_k = \neg m_i  \lor \neg m_j  \nonumber \\
. \nonumber 
   \omega_k   = \omega_i \cup [\omega_j  \cap \omega_p] &\Longleftrightarrow m_k = m_i  \lor [m_j \land m_p ]\nonumber \\
   \ldots \nonumber
\end{align*}

Where $[]^C$ stands for the set complementation operation respectively to $S_E$.
Observe also that to each set operation does correspond a logical manipulation of models, and vice-versa.

Table~\ref{tab:opers} presents the 16 possible logical operations between two logical variables $m_i$ and $m_j$ 
yielding $m_k$, but out of them 4 are not really useful for obtaining new combinations of models 
(and datasets): $m_k = \text{TRUE}$, $m_k = \text{FALSE}$, $m_k = m_i$ and $m_k = m_j$.

\begin{table*}[h!]
\centering
\renewcommand{\arraystretch}{1}
\begin{tabular}{| c | c || c | c | c | c | c | c | c | c | c | c | c | c | c | c | c | c ||}  
\hline
$X$  & $Y$ &  $op_0$ & $op_1$ & $op_2$ & $op_3$ & $op_4$ & $op_5$ & $op_6$ & $op_7$ & $op_8$ & $op_9$ & $op_{10}$ & $op_{11}$ & $op_{12}$ & $op_{13}$ & $op_{14}$ & $op_{15}$  \\
\hline  \hline
0  & 0  & 0 & 0 & 0 & 0 & 0  &  0 & 0 & 0  & 1  & 1  & 1 & 1 & 1 & 1 & 1 & 1 \\  \hline 
0  & 1   & 0 & 0 & 0 & 0 & 1  &  1 & 1 & 1 & 0  & 0  & 0 & 0 & 1 & 1 & 1 & 1\\ \hline 
1  & 0   & 0 &0 & 1 & 1 & 0  &  0 & 1& 1 & 0  & 0  & 1 & 1 & 0 & 0 & 1 & 1\\ \hline 
1  & 1   & 0 &1 & 0 & 1 & 0  &  1 & 0 & 1 & 0  & 1  & 0 & 1 & 0 & 1 & 0 & 1\\
\hline
\hline  
\end{tabular}
\renewcommand{\arraystretch}{1}
\caption{The 16 possible logical operations between two logical 
(or Boolean) variables $X$ and $Y$.  Remarkably,
some of them --- such as ``not'', ``and'' , and ``or'' --- are closer to human cognition
in the sense of being more frequently employed.  Provided the conditions for
the bijective association between datasets and models is fulfilled, there will be
a set operation respective to each of the 16 logical operations.}\label{tab:opers}
\end{table*}

It is also interesting to observe that it is possible to implement every logical operation
in terms of $\neg (x \lor y)$ or $\neg (x \land y)$, among other possibilities,
which in the data domain becomes $[x \cup y]^C$ and $[x \cap y]^C$, respectively.

It can be shown that there exists a total of $2^n$ possible logical operations between
$n$ logical variables.  As a consequence, the number of possible cases steeply
increases with the hierarchy of the sets or models.  However, the currently available
extensive computational resources
can be applied, possibly incorporating optimization techniques.  At the same time,
the incorporation of many hierarchical levels in the description of a newly obtained
model also implies that description to become less tangible by human perception,
and therefore more \emph{complex} and \emph{abstract}.
For these reasons, the efficiency of the modeling framework greatly depends on the
choice of models for the initial framework, in the sense that some of these choices
may contribute to explaining a new dataset in terms of a relatively (or, ideally, minimal)
combination of the previous models.  

The subject of obtaining new models through set operations between the existing
datasets (or logical between models) is as extensive as it is interesting and cannot
not be fully addressed here.  However, an interesting approach consists of employing
intersection, union, complementation and difference between a small number of
datasets.

When translated to the human perspective,
this intrinsic combinatorial complexity of possible models becomes closely related to
the concept of complexity, because it becomes more and more expensive~\cite{CostaComplex} to 
develop and understand highly hierarchical models.  This provides a motivation for
having experts in specific areas, who can substantially contribute to integrating other models
through collaborative exchanges.

It is also interesting to observe that, instead of understanding that a model
needs to be satisfied by every data element of the respective dataset, it would also
be possible also to allow the validity of the model not to be restricted to 0 (false)
or 1 (true), but to depend on a graded merit figure such as the number of elements
in the dataset that satisfy the model. The immediate implication of this is the loss
of the bijective association between datasets and models.  Now, instead of
being underlain by a formal logic consistency, the modeling approach starts being
understood as an optimization problem.

By blurring the frontier between data and models, the $M^*$ approach paves the way
to  several interesting possibilities, 
including the definition of a paired algebra of dataset and models queries and manipulations.
By `algebra' we mean the ability to represent datasets or models as variables, or symbols
that can be solved or interrelated through set operations (in the case of datasets) or logical equations
(models).    The association of a new dataset to a model corresponding to a combination 
of the previously available models provides not only a way to account for the respective dataset,
but its intrinsic logical construction can also provide insights for computationally
to decide if specific data elements satisfies that respective model. In addition, the obtained
models may also provide indications about how the datasets were generated, sampled and obtained.

It is important to stress that the combinations of the existing datasets while searching
for a match needs to be preceded by including in every existing dataset associated to
a respective model all the elements in the new dataset that satisfies that respective model,
and revising the overall consistency of the new state of the modeling framework.
It is also possible to proceed gradually from new single data elements to the whole
new dataset by progressively considering subsets of the latter having increasing
sizes.  For simplicity's sake, the present work will be restricted to updating the new
individual elements and then the considering the whole new dataset.

Another possibility accounted for by the proposed framework consists in performing
logical combinations between the model statements and then seeking which among the
existing datasets can satisfy a new modeling statement.   This same mechanism immediately
provides the means for building programs for obtaining specific properties (determined
by the model) to characterize and analyze respective datasets.

The above two possible approaches aiming at combining datasets or combining models are
henceforth understood as being \emph{data-driven} and \emph{model driven}.  These
two ways of integrating information and knowledge seem to correspond to the main
manners in which humans perform these two important activities.  A third possibility consists
of evaluating the pairs of datasets and models from the perspective of the current modeling
framework.

As an example, consider that the original dataset $\omega_4$ in Figure~\ref{fig:model} becomes
important enough to motivate the development of a respective model.  We can approach the
solution to this essential problem by searching for a combination between the datasets already
explained by models that is identical to $\omega_4$.   This can be done by checking between
the result of set operations between the already instantiated datasets, such as:
\begin{equation}  \label{eq:ex1}
   ?\omega_i  \; \cap \; ?\omega_j \equiv \omega_4
\end{equation}

Sought datasets are henceforth represented with a preceding question mark, i.e.~$?\omega$, 
while the $\equiv$ symbol stands for being equivalent or identical. 

In case it is verified that the datasets $\omega_2$ and $\omega_7$ satisfy Equation~\ref{eq:ex1},
by using set intersection, we can immediately derive  the model of $\omega_4$ as necessarily
corresponding to:
\begin{equation}  
   m_4 = m_2 \land m_7
\end{equation}

\begin{figure}[h!]  
\begin{center}
   \includegraphics[width=0.9\linewidth]{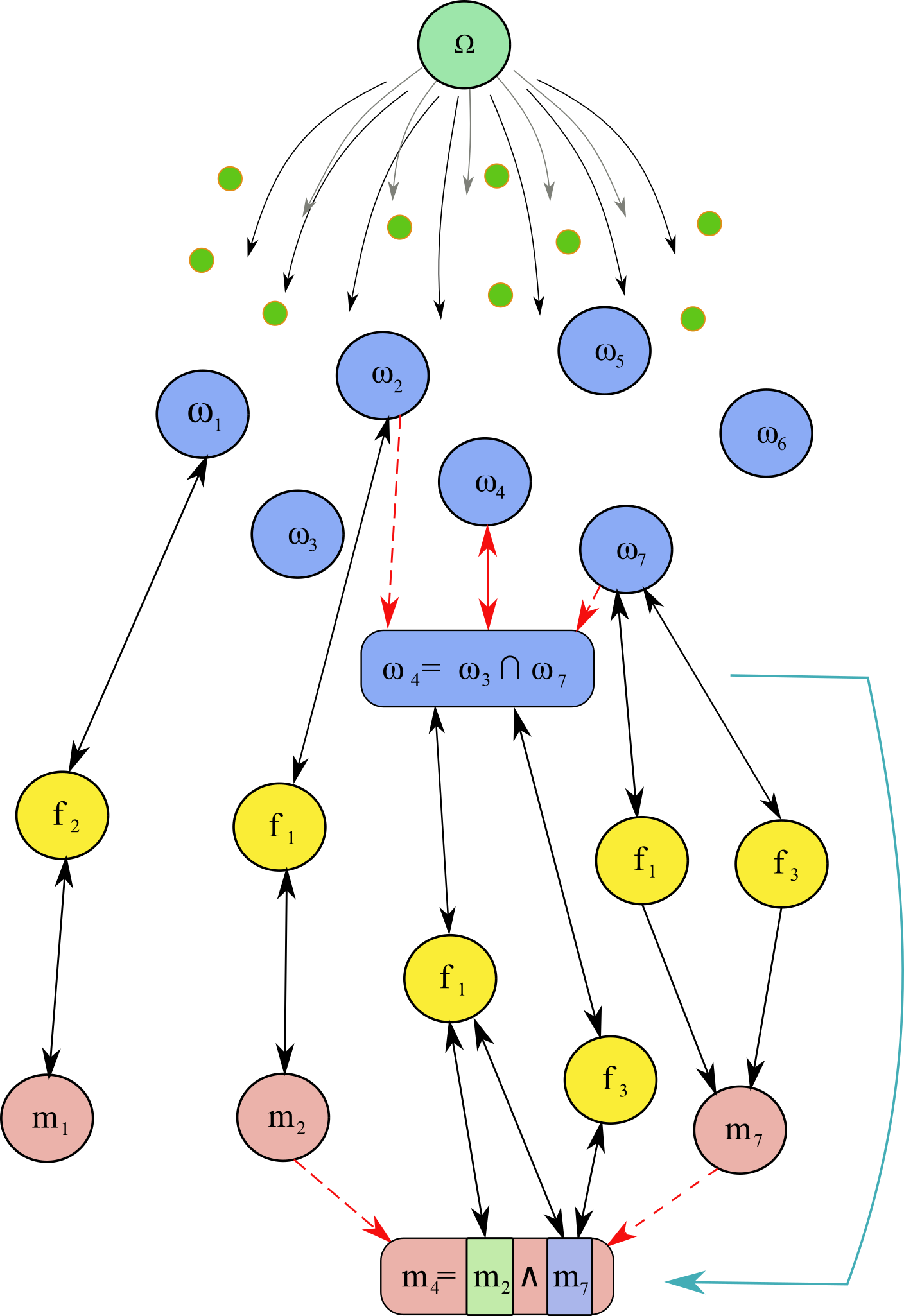}  
    \caption{A new model is born.  Having found that the dataset $\omega_4$ corresponds
    identically to the intersection between $\omega_2$ and $\omega_7$, and since both
    these two datasets are associated to respective models,  it becomes possible
    to derive a model that satisfies the data elements in $\omega_4$.  This can be done
    by defining the new model $m_4$ as having as input the features inherited from both
    $\omega_2$ and $\omega_7$, which correspond to $f_1$ and $f_3$.  The model
    $m_4$ is immediately provided by the logical and operation between the models
    $m_2$ and $m_7$.}
    \label{fig:oper_model}
    \end{center}
\end{figure}

The above described development of a model for the dataset $\omega_4$ in Figure~\ref{fig:model}
is illustrated in Figure~\ref{fig:oper_model}.
 
The developed framework also allows the features of $m_4$ to be immediately inherited 
from the features originally associated to $m_2$ and $m_7$.  As feature $f_1$ and $f_3$ 
were required by both the original models, they also become pre-requisites of the new model
$m_4$.  Observe that the different logical components of the latter model may receive different
sets of features as input, as is the case for the present example.  At the same time, it 
should be kept in mind that the bijective association between datasets and models depends
critically on the choice of features as well as the current dataset and modeling framework.

Another example of the possibilities allowed by the suggested approach concerns algebraic
equations as:
\begin{equation}  
   ?m_i  \; \cup \; ?m_j \equiv m_k
\end{equation}

In other words, we would search in the current model framework $M$ for a pair of models
satisfying the above condition.  

It is also possible to derive \emph{hybrid} algebraic equations such as:
\begin{equation}  
   ?\omega_i  \; \cup \; \omega_j \equiv m_k \; \land  \; ? m_p
\end{equation}

This type of equation can be solved either by translating the dataset-based side
(lefthand) into the respective model logical equation, or vice versa, and then applying the
above mentioned procedures.  In the former case, we would have:
\begin{equation}  
   ?\omega_i  \; \cup \; \left[\omega_j \right]^C \equiv m_k \; \land \;? \neg m_p   \Longleftrightarrow ?m_i  \; \land \; \neg m_j \equiv m_k \; \land \; ? \neg m_p \nonumber
\end{equation}
 
Alternatively, we could make:
\begin{equation}  
   ?\omega_i  \; \cup \; \left[\omega_j \right]^C \equiv m_k \; \land \;? \neg m_p   \Longleftrightarrow ?\omega_i  \; \land \; \left[\omega_j \right]^C  \equiv \omega_k \; \cap \; ? [\omega_p]^C \nonumber
\end{equation}
 
\begin{figure}[h!]  
\begin{center}
   \includegraphics[width=0.9\linewidth]{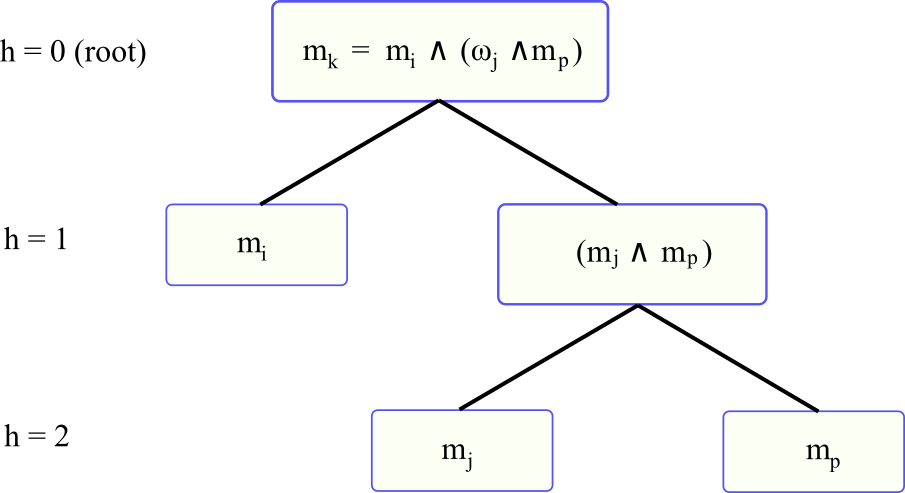}  
    \caption{As models are developed and integrated by the proposed methodology, a
    respective hierarchy is progressively established as illustrated in this hypothetical example.  
    The hierarchical levels are indicated as $h$, with $h=0$ corresponding to the
    root of the tree representing the hierarchy.}
    \label{fig:hierarchy}
    \end{center}
\end{figure}

Interestingly, as models (as well as the respectively associated datasets)
are progressively combined, a respective hierarchy is defined, as
illustrated in Figure~\ref{fig:hierarchy}, which corresponds to the the last of the models
shown in the list above.

\begin{figure}[h!]  
\begin{center}
   \includegraphics[width=0.9\linewidth]{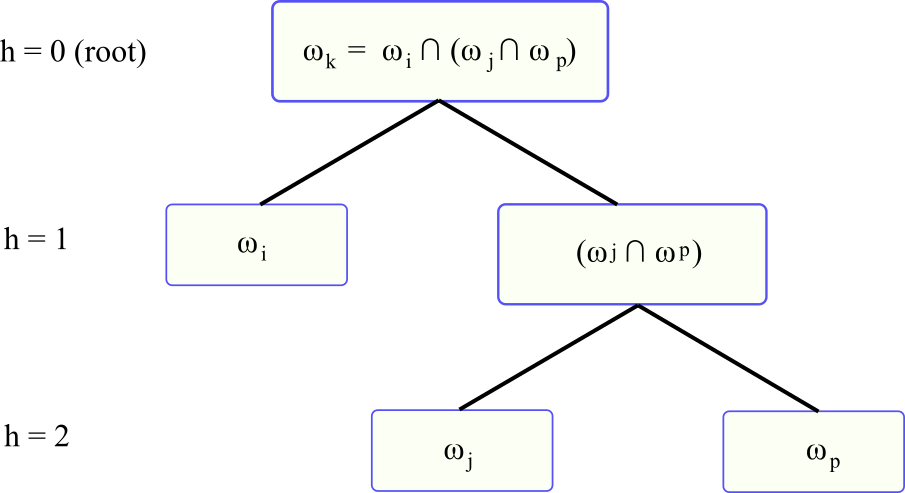}  
    \caption{The dataset hierarchy respectively associated to the model hierarchy in 
    Fig.~\ref{fig:hierarchy} as a consequence of the bijective
    association between datasets and models established by the
    $M^*$ approach.}
    \label{fig:hierarchy_data}
    \end{center}
\end{figure}

As a consequence of the bijective association between datasets and models
established by the $M^*$ model, we immediately have that the above model
hierarchy to be respectively reflected into the data hierarchy illustrated in 
Figure~\ref{fig:hierarchy_data}.

Provided the number of datasets associated to respective models is not too large and that
we are not aiming at several hierarchical compositions of models, it is
possible to search for the solution of problems by systematically checking every possible
combination of the set/logic operations  while progressively increasing the number of datasets or
models.

It is interesting to observe that the level of \emph{abstraction} in the modeling framework
increases as we move from the leaves to the root of the respective tree.  That is so because
the understanding of composite models demands the understanding of the preceding models.
At the same time, the level of \emph{generalization} may increase, in case the combinations
involve the union of sets, or decrease, as implied by intersections between the sets.  As such,
the proposed framework accounts for these two important paradigms, while also indicating
that the abstraction increases in both these cases, as it is ultimately related to the complexity
of the respective logical model.

The proposed $M^*$ approach also relates to the critically important concept of
\emph{causality}.  It is posited here that, at least as typically understood by humans,
causality corresponds precisely to the bijective association established between
a dataset of relevance and its respective model, and more specifically in the fact that
\emph{every} element of a data set satisfies, implies (or causes) the model.  In other
words, it is only the full presence of all the conditions of a model that can 
enable the model verification.  Incidentally, observe that this possible definition of
causality, which is after all a human concept just like complexity, implies the
time sequentiality that is often used to characterize causality.  After all,
having all the data elements in the dataset representing the event of interest
triggers the model (decision) conditions to be satisfied.  The
situations corresponding to incorrect identifications of causality would correspond to
models only partially associated to an incorrect, though related model, which 
will otherwise lead to correlations between the observed events.  Recall that
in the proposed framework a same model may satisfy (imply or cause) two or more 
models.

In cases where the number of data elements satisfying each model is taken as 
a graded indication of model adherence, the existence of crossed connections,
i.e.~a data element satisfying more than one model, implies in respective
correlations between the activations of models that are related to the datasets
in a non-necessarily causal manner.

\section{Case-Example: Binary Lattices}

As a more concrete example of representing the construction of a model framework
by using the concepts and techniques suggested so far, we consider each data
element $x_i$ to correspond to each of the possible instances of a binary lattice
or array with dimension $L \times L$.  By `binary' it is meant that the lattice elements
can only assume the values `0' or `1'.  

Therefore, we have $\Omega$ to correspond to every possible binary pattern on a lattice,
while the datasets will correspond to subsets of the power set of $\Omega$.

\begin{figure}[h!]  
\begin{center}
   \includegraphics[width=0.7\linewidth]{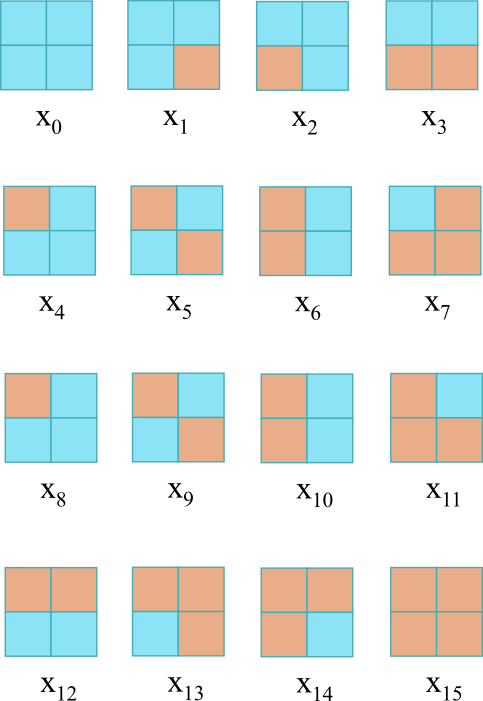}  
    \caption{All the possible $2^{(2^2)} = 16$ data elements that can be derived from a binary lattice
    with dimension $2 \times 2$.  Zeros can be understood to correspond to the
    blue points, and ones to the brown points.  }
    \label{fig:lattice}
    \end{center}
\end{figure}

Given an integer value $L$, a total of $N_L = 2^{(L^2)}$ possible data elements are respectively
defined.  Figure~\ref{fig:lattice} presents the set of all possible patterns for $N = 2$.  

Observe that the number of patterns increases in a very steep, exponential manner. 
We shall adopt $L=3$ for our first case example, therefore implying $N_L = 512$
possible data elements or basic patterns in $\Omega$.

For simplicity's sake, we will assume that $E$ contains all possible data elements of $\Omega$,
but this is not a necessary requisite for the $M^*$ approach.

Let's assume that several datasets have already been singled out for their potential
relevance and associated to respective models, as presented in Table~\ref{tab:props}.

\begin{table*}[h]
\renewcommand{\arraystretch}{1}
\begin{tabular}{|| c |c | c | c || c  |}  
\hline
$m_i$ & \emph{ textual description of the model}  & \emph{features/structures} & \emph{decision}  &  \emph{size} \\
\hline  \hline
$m_1$ & contains only an isolated point&  number of points $n$  &   $n = 1$ & 9 \\  \hline
$m_2$ & contains $2$ points  &  number of points $n$  &   $n = 2$   & 36   \\  \hline
$m_3$ & contains $3$ points  &  number of points $n$  &   $n = 3$  & 84    \\  \hline
$m_4$ & contains $4$ points  &  number of points $n$  &   $n = 4$   & 126  \\  \hline
$m_5$ & contains at least $3$ points  &  number of points $n$  &   $n \geq 3$  & 466  \\  \hline
$m_6$ &  conn. comp. contains $\left[1,1\right]$  & conn. comp.  $\kappa_k$  &   $\left[1,1\right] \in \left\{ 
\kappa_k \right\}$  & 256  \\  \hline
$m_7$ &  conn. comp. includes  $\left[1,1\right]$ and $\left[N,N\right]$  &  conn. comp. $\kappa_k$  &  $\left[1,1\right], \left[N,N\right] \in \left\{ \kappa_k \right\}$   & 88 \\ \hline
$m_8$ & the foreground set is thin  &  width $w$ at each point  &   $w = 1$ or 2 for every point   & 291    \\  \hline
$m_9$ & opposite of $m_8$ &  width $w$ at each point  &   $w  > 1$ for at least one point & 221    \\  \hline
\hline  
\end{tabular}
\renewcommand{\arraystretch}{1}
\caption{A possible model framework $M$ for the binary lattice case-example.}\label{tab:props}
\end{table*}

This table includes the textual meaning of each model, the involved features and data
structures, the formal condition representing the model, as well as the size of the dataset
respectively satisfying each model.  Recall that each of these models has a respectively associated
dataset.  Also, it should be realized that the choice of this initial modeling framework is, in
principle, completely arbitrary, though the effective combination of models will also depend
critically on these initial choices.
 
For simplicity's sake, each of the binary lattice elements, henceforth called a \emph{point},
is here understood to be a square with four margins, two vertical and two horizontal.  The 
points with value `0' are called \emph{background} points while those equal to `1' are said to be \emph{foreground} points.
A point $a$ in the binary lattice is said to be \emph{adjacent} to another point $b$ provided
they share a vertical or horizontal margin.  A \emph{connected component} is a set of foreground 
points so that it is possible to move between any pair of their constituent points through 
adjacent margins.
The \emph{local width} of a set of foreground points is henceforth understood as the number of 
adjacent foreground points, i.e.~neighbors.    A binary lattice element is said to be thin
whenever each of its foreground points has width 1 or 2 (e.g.~\cite{shapebook}).

The models adopted in this example framework range from being very simple (e.g.~$m_1$)
to moderately complex (e.g.~$m_6, m_7, m_8$), though this is a largely subjective
classification.  It is interesting to observe that us, humans,
tend to have more difficulty in handling the complement of a model than its direct definition, as is the
case $<9>$.

\begin{figure}[h!]  
\begin{center}
   \includegraphics[width=0.7\linewidth]{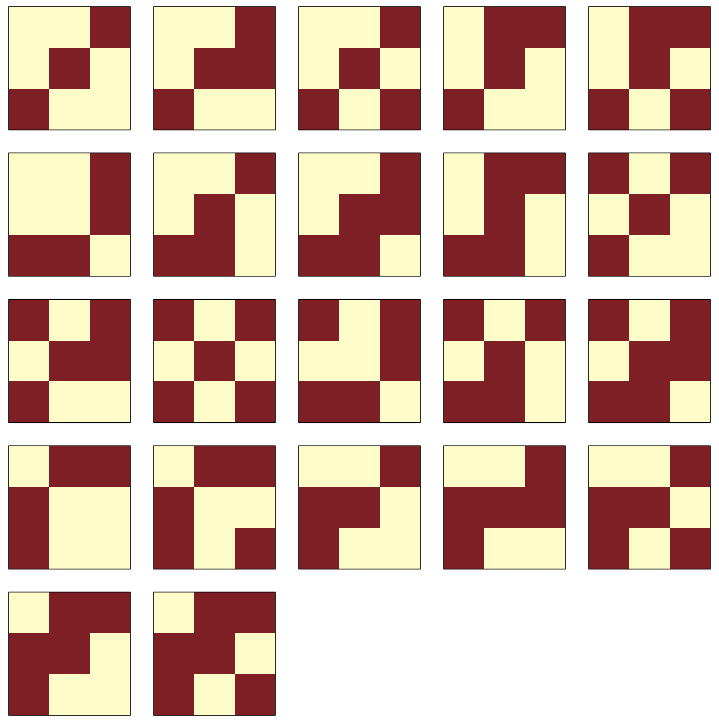}  
    \caption{A puzzle. A dataset $\omega_{10}$ containing possible data elements derived from the
    $3 \times 3$ binary lattice needs to be associated to a model.  Can it be derived from the
    model framework in Table~\ref{tab:props}?  What is the property shared by these patterns?}
    \label{fig:puzzle}
    \end{center}
\end{figure}

Let's now proceed to the dataset $\omega_{10}$ shown in Figure~\ref{fig:puzzle}.
 
What is the model fully explaining this dataset?  Can it be derived from the modeling framework
in Table~\ref{tab:props}?  This problem can be tackled by considering the several types of
combinations of the existing datasets satisfying algebraic constructions such as:
\begin{eqnarray}  
   ?\omega_i  \; \cup \; ?\omega_j \equiv \omega_{10} \nonumber \\
   ?\omega_i  \; \cap \; ?\omega_j \equiv \omega_{10} \nonumber \\
   ?\omega_i  \; - \; ?\omega_j \equiv \omega_{10} \nonumber \\
   ?\omega_i  \; \cup \; ? \left[\omega_j \right]^C \equiv \omega_{10} \nonumber \\
   ?[\omega_i]^C  \; \cup \; ? \omega_j \equiv \omega_{10} \nonumber \\
   \ldots \nonumber \\
   \left[ ?m_i  \; \cup \; ?m_j \right]^C \equiv \omega_{10} \nonumber \\
   \left[?m_i  \; \cap \; ?m_j \right]^C \equiv \omega_{10} \nonumber \\
   (?m_i  \; \cup \; ?m_j) \; \cup \; ?m_k  \equiv \omega_{10} \nonumber \\
   \ldots \nonumber
\end{eqnarray}

As it happens, it can be verified that the dataset of interest can be obtained through the 
combination $m_7  \; \cap \; m_8$ is identical to $\omega_{10}$, therefore implying
the following respective model:
\begin{equation}  
    \omega_{10} \equiv \omega_7  \; \cap \; \omega_8   \Longleftrightarrow  m_{10} \equiv m_7 \land m_8 \nonumber
\end{equation}

It follows that, as the number of existing models (and datasets) increases, the higher the probability
of finding a combination of those models that can explain a new dataset drawn from $\Omega$.  
Recall that the compositions between
the available models give rise to a respective hierarchical organization.  Also, a new dataset
can be explained by a completely new model not directly related to the existing ones, though
perhaps sharing some of the adopted features.

Another interesting situation arises when a new model $m$ is given regarding whether it will
satisfy any of the existing dataset.  This problem can be approached by trying to
identify a logical combinations between the existing models that yield the new model,
or by checking every existing dataset against the new model.

For instance, let the new model $m_11$ be textually defined as ``the dataset contains the
shortest connected component comprising both the lattice elements $[1,1] \emph{ and } [N,N]$.''
A possible first step is to try to translate this condition in terms of the available measurements
and models.  First, we select model $<7>$, because it selects all data elements that contain
at least a connected component containing $[1,1] \emph{ and } [N,N]$.  Then, we take into
account that the shortest possible path necessarily contains 3 points, which can be verified
from model $<2>$ while making $P=3$.  The sough model then can be obtained as:
\begin{equation}  
    m_{11} = m_2 \land m_7 \nonumber
\end{equation}

The dataset satisfying this condition can be immediately identified as corresponding to the
dataset associated to both $\omega_2$ and $\omega_7$, corresponding to the 3-point diagonal
between the points $[1,1] \emph{ and } [N,N]$.  Observe that the logical construction of the
obtained combined model allows us to directly obtain, through logic-computational means,
the respective means for verifying the adherence of specific data elements.

\section{The Meta Model for Incomplete Data}\label{sec:epsilon}

So far, we have understood that all elements in all datasets satisfy the respectively
associated model.  However, it is possible that one or more elements currently
in a given dataset do not satisfy the respective model.  Possible causes for this
include errors in the features determination, model inconsistencies  implying the mathematical
implementation not to correspond to the textual characterization of the model,
or errors in storing and handling the datasets and/or models.   These situations
will be henceforth understood as corresponding to the presence of error or noise.

From the perspective of the present work, the most important consequence of errors
and sampling is the loss of the bijective association that is critical for the consistency of the
$M^*$ reference model, which leads to modeling, decision and classification errors.

In this section we describe an adaptation of the $M^*$ model, here called
$M^{<\epsilon>}$, which can be considered for dealing with the above characterized
situations, as well as for applications where only approximate verification of the
conditions implied by the models are allowed.   The underlying idea in all these cases
is to adopt some effective means for quantifying the similarity between any two sets.

A possibility to cope with errors and sampling would be to relax the binary
decision on the validity of a model that is characteristic of the $M^*$ structure.
This could be done by having by grading the degree of validity of a model.

In the present work, we will address the sampling and error limitations by the adoption of
the following index that can express the similarity of two discrete sets $A$ and $B$:
\begin{equation}
   \Lambda(A,B) = \frac{|A \cap B|}{|A \cup B|}
\end{equation}

where the operator $| A |$ stands for the cardinality (or number of elements) of the 
set $A$.

It can be verified that, conveniently, the above index is intrinsically normalized
as $0 \leq \Lambda \leq 1$, so that it does not depend on the size of the sets.

Observe that the above mentioned graded validity of models can also be combined 
with the adoption of the $\Lambda$ similarity for deciding on the associations between 
datasets.

As an example, let a set $A = \left\{1, 2, 3, 4, 5, 6, 7 \right\}$, so that $|A| = 7$.
Suppose a new set $B = \left\{1, 2, 3, 4, 5, 6, 7, 8\right\}$ is to be compared with
set $A$.  This situation could be implied by obtaining a new version of a previous
dataset, but incorporating by mistake the element `8'.

In this case, we would have $A \cap B = \left\{ 1, 2, 3, 4, 5, 6, 7 \right\}$
and $A \cup B = \left\{ 1, 2, 3, 4, 5, 6, 7, 8 \right\}$, therefore implying $|A \cap B| = 7$
and $|A \cup B| = 8$, from which we obtain:
\begin{equation}
   \Lambda(A,B) = \frac{|A \cap B|}{|A \cup B|} = \frac{7}{8} = 0.875
\end{equation}

The index $\Lambda()$ therefore provides an interesting resource for checking if a new dataset could
correspond to a noisy version of any of the existing datasets.  This can be done by
adopting a threshold $T$, and discarding  any new dataset $\tilde{\omega}$ for
which $\Lambda(\tilde{\omega},\omega_i) \leq T$ for any of the existing sets $\omega_i$.

Let us now illustrate the situation where a new version $B$ of the existing set $A$ is
obtained while overlooking some elements, e.g.~$B = \left\{  1, 2, 3, 5, 7 \right\}$.
In this case, we would get:
\begin{equation}
   \Lambda(A,B) = \frac{|A \cap B|}{|A \cup B|} = \frac{5}{7} = 0.7142
\end{equation}

As the obtained value is relatively high, it would suggest that the set $B$ does not correspond
to a new dataset and therefore can be merged into $A$ or understood as being the same
model.

However, in case $B = \left\{  1, 5, 10, 12, 20 \right\}$, we would obtain:
\begin{equation}
   \Lambda(A,B) = \frac{|A \cap B|}{|A \cup B|} = \frac{2}{7} = 0.2857
\end{equation}

which is much smaller than in the previous case, suggesting that the set $B$ does
correspond to a new dataset.

As it will be presented in Section~\ref{sec:prob}, the above similarity index can
be adapted to datasets associated to respective probabilistic densities.

It should be observed that there are several other possible indices and methodologies
that can be applied to deal with error and noise influencing data, features, and models.
However, the above described alternative provides a particularly interesting approach especially
given its conceptual and computational simplicity.  In addition, though we currently
discussed only possibilities for trying to avoid the incorporation of incorrect datasets,
there are many other implications of incorrect or missing data and modeling that deserve
to be further addressed at more length.

In brief, the $M^{<\epsilon>}$ meta model, as described here, can be simply understood
as the $M^*$ model that uses the adopted similarity index in order to identify the most
likely combination of existing models while explaining a new dataset as well as to 
decide whether two datasets could be treated as being the same.

Also important to realize is that the above discussed errors and sampling 
imply in loosing the bijective association which is required for consistency
in the reference $M^*$ model, implying in respective modeling errors, 
classifications, and decisions.

\section{Case Example:   Elementary Number Theory}

In order to illustrate the potential of the $M^{<\epsilon>}$ approach, a model of
numeric sets  taking into account the property of a number being a multiple of 
some radix is described in this section.  This example involves new datasets that cannot
be exactly explained by any of the models in the current modeling framework.

We start by defining $\Omega = \left\{ 2, 3, 4, \ldots, 20\right\}$.  The number 1
is omitted as it is a trivial divisor of any natural number.  

The datasets $\omega_i$, $i= 1, 2, \ldots, 19$ will correspond to the multiples of 
$i+1$ up to $20$.   Thus, $\omega_3 = \left\{ 3, 6, 9, 12, 15, 18 \right\}$.  The respective
models are immediately derived from the respective multiplicity property.  For instance,
$M_3$ corresponds to ``all the numbers smaller than 20 that are divisible by 3''. 
Therefore, the adopted modeling framework contains a total number of 19 pairs $(\omega_i,m_i)$.

The first important point to be taken into account that these 19 models are by no means
sufficient for explaining most of the possible new datasets that can be drawn from $\Omega$.
However, as we will see, the adoption of the $\Lambda$ similarity allows a surprisingly
good performance while being capable of providing interesting insights about possible
explanations an interrelationships, even if no perfect combination can be found.

A modeling engine was implemented, using list manipulations in R, considering the following 
set operations ($A$ and $B$  are any of the existing datasets) shown together with the
respective logical operations:

\begin{itemize}
\item $A  \Longleftrightarrow m_A$
\item $A ^C \Longleftrightarrow \neg m_A $  
\item $A \cap B \Longleftrightarrow m_A \land m_B$    
\item $(A \cap B )^C = A^C \cup B^C \Longleftrightarrow \neg(m_A \land m_B)$   (De Morgan)
\item $A ^C \cap B = B - A\Longleftrightarrow \neg m_A \land m_B$
\item $A \cap B ^C = A - B \Longleftrightarrow m_A \land \neg m_B$
\item $A \cup B \Longleftrightarrow m_A \lor m_B$ 
\item $(A \cup B )^C = A^C \cap B^C \Longleftrightarrow \neg(m_A \lor m_B)$  (De Morgan)
\item $A ^C \cup B \Longleftrightarrow \neg m_A \lor m_B$
\item $A \cup B ^C \Longleftrightarrow m_A \lor \neg m_B$
\item $(A^C \cap B) \cup (A \cap B^C) \Longleftrightarrow m_A \oplus m_B$ (xor)
\item $(A \cap C) \cup (A^C \cap B^C) \Longleftrightarrow  m_A \odot m_B$   (xnor)
\end{itemize}

Let's now consider the data-driven query relative to the new dataset $\omega = \left\{ 
2, 4, 6, 8, 10, 12, 14, 3, 6, 9, 12, 15\right\}$.

Only operations between two Boolean variables are considered for the sake of
simplicity and also for keeping the results more accessible to human interpretation.

The engine found  $\Lambda = 0.769$, respective to the model 
$M = M_2 \lor M_3$, which corresponds to the union of the multiples of 2 and 3.
Though the similiarity index is not maximum, the provided explanation is still quite reasonable
even if the given dataset cannot be fully expressed by the obtained combination
(there are some values missing in $\omega$).

Let's consider now another example, respective to the new dataset $\omega = 
\left\{ 8, 10, 12, 14  \right\}$.  We get $\Lambda = 0.5$ for two approximate solutions:
$\omega_{12} \cup \omega_{14} $  and $\left( \omega_4 \cap (\omega_{10})^C \right)
\cup \left( (\omega_4)^C \cap \omega_{10} \right)$.  Observe that, though
all the numbers in this given dataset are multiples of 2, the set only contains 4
out of the 10 elements in $\omega_2$, so the result obtained is still fully compatible.
Given that $\Lambda=0.5$ can be considered too low, a new model would need to
be defined for this dataset, as it cannot be approximated by combinations of those
in the existing modeling framework.  In this case, as observed in Section~\ref{sec:M_star},
it is also possible to associate to a restricted version of $\omega_2$, 
i.e.~$\omega$ to $\omega_2 - (\omega_2 - \omega)$.  Another possibility is to
take into account a further feature, such as being comprised within a given
minimum and maximum values.

Now, let's make $\omega = \left\{ 2, 3, 5, 7, 11, 13, 17, 19 \right\}$.  The result
provided by the engine is $\Lambda = 0.875$, corresponding to two equally
similar approximated solutions $(\omega_2 \cup \omega_9)^C$ as well as 
$\left( \omega_2 \cap \omega_{15} \right) \cup \left( (\omega_2)^C \cap (\omega_{15})^C \right)$.
Interestingly, the prime numbers in the range from 2 to 20 could be well modeled
in terms of the two combinations of sets, with a relatively high similarity index.

As another example, let's consider $\omega = \left\{ 3 \right\}$.  In this case we get
$\Lambda = 1/3$, with 5 respective possible approximated combinations.  This example
illustrates the issue that it is difficult to express a small set as a pairwise combination
of larger sets such as most of those in the existing modeling framework.

Let's now have the dataset $\omega = \left\{ 4, 8, 12, 16, 20\right\}$.
By querying the described engine, we have two solutions for $\Lambda =1$:
$\omega_4$ and $\omega_2 \cap \omega_4$, therefore capturing the fact that
this last example dataset contains all elements between 1 and 20 that are both multiple of 2 and 4.

As a last example, we have $\omega = \left\{ 1, 3, 5, 7, 9, 11, 13, 15, 17, 19 \right\}$,
which corresponds to the odd numbers between 2 and 20.  The exact
solution $[\omega_2]^C$ is provided by the engine.

As above illustrated, even though not ensuring full accuracy, the application of the 
$M^{<\epsilon>}$ can still provide valuable insights while understanding datasets and
trying to find models for them.

\section{The $M^{\left<\sigma\right>}$ Stochastic Meta Model }\label{sec:prob}

There are several abstract and real-world situations in which the datasets are characterized
by respective features that may extend continuously along the respective axes in the
feature space.  Or, more importantly, there are cases in which $\Omega$ contains a huge
number of elements, which tends to be the case for several real-world situations
(e.g.~the set of all possible butterflies).  Mathematically, the feature-based representation
of these sets can be properly obtained in terms of respective multivariate 
\emph{probability densities} representing data elements distribution
in the adopted feature space.  At the same time as this probabilistic approach enables
the consideration of many interesting problems, it also implies that the full consistency
between data and model that is characteristic of the $M^*$ to be undermined.
This is to a great extent a consequence of the fact that the probability densities
associated to specific models/categories often overlap one another.  Each probability
density is intrinsically associated to a respective \emph{random variable}, or 
measurement.

This type of representation can be shown to provide virtually every statistical information
that may be required regarding the dataset as described by the adopted features.
For instance, the probability of observing all the data elements contained in a given
subset of the feature space can be estimated in terms of the hyper-volume of the
density taken on that region.  The reader should not be put off by the seeming sophisticated
adopted mathematical concepts, as the overall idea and principles are likely to be
grasped with the help of the case-example provided in Section~\ref{sec:case2}.

The representation of feature-based discrete, sampled datasets also
leads to the possibility of applying Bayesian decision (e.g.~\cite{shapebook,Koutrombas,DudaHart}) 
in order to decide what is the
most likely category given a specific data element.  This same approach also provides
subsidies for estimating the probability of making incorrect decision.  Even more
importantly, the above outlined Bayesian decision method can be show to provide
optimal results in the sense of minimizing the chances of making decision errors,
However, this important property requires the availability of \emph{exact} probability
densities, but good results should be obtained for representative samples of data.

Because it is impossible to obtain an infinite number of samples
allowing the complete characterization of these continuous variables, we need to resource
to some suitable methodology capable of yielding satisfactory estimations in terms of
\emph{estimated} probability densities.   The integration of this approach into the
suggested $M^ *$ meta model leads to the $M^{<\sigma>}$ variant, capable of
addressing situations characterized by incomplete data sampling.  The remainder of
this section presents a description of this approach.

Let $\omega_i$, $i = 1, 2, N_{\omega}$, be datasets sampled from a universe
$\Omega$.  Each of these datasets $\omega_i$ is chosen to be characterized
in terms of a set of random variables (or features) $f_j$,  $j = 1, 2, N_{f}$.  In order to obtain a suitable
probability density representing each of these datasets, it is possible to perform a
kernel expansion (e.g.\cite{DudaHart,Koutrombas,CostaConv,kernel:wiki} 
on that set, with each individual data element being represented as
a Dirac's delta function $\delta(\vec{f})$, 
$\vec{f} = \left[ f_1 \; f_2 \; \ldots \; f_{N{_f}} \right]$.  

The gaussian kernel represents
an interesting choice given its mathematical properties, but other kernels can be more
suitable depending of the type of datasets and features of their original probability
densities.  The expansion itself can be performed by convolving 
(e.g.~\cite{CostaConv,brigham:1988}) the given dataset
with a normalized version of the kernel.  The estimated probability density 
obtained by kernel expansion  of each dataset $\omega_i$ is henceforth expressed 
as $p_i (\vec{f})$.

Now, in order that each dataset be associated to a \emph{support region} having finite hyper-area,
we perform a thresholding operation on the respective estimated probability density.
The resulting support hyper-region is henceforth indicated as $\rho_i$, which thus is
necessarily a subset of the respective feature space.

The value of the threshold can be determined as corresponding to the situation in which
the respectively obtained support region contains a fixed ratio $\chi$ of the possible
elements (this is related to the concept of percentile).  In practice, one is likely to choose
large values of $\chi$ in order to retain a representative set, but other approaches may 
also be of interest. 

Observe that the option of defining the support regions in terms of percentiles implies
a more sparse dataset to cover a larger area than a more compact counterpart.
These situations may be addressed by taking into account also the density values
associated to the support regions, reflecting the fact that the less dense portions of the 
support region will have smaller weight.

It is also necessary to derive the overall probability density as defined by all elements
in all the existing datasets.  This can be done by first obtaining the union of all
available datasets, i.e.~ $\Gamma = \omega_1 \cup \omega_2 \cup \ldots \cup
\omega_{N_{\omega}}$ and then performing a kernel expansion possibly considering
the same $\chi$ as adopted for estimating the other support regions.  The estimated
overall probability density is henceforth represented as $p(\vec{f})$.

The derivation of the support regions for each involved dataset has the immediate
benefit of having finite area, immediately allowing them to be combined respectively
to the set operations involved in the described $M^*$ modeling approach.  

We are now in a position to generalize the similarity index described in Section~\ref{sec:epsilon} 
in order to quantify the similarity between two sets $A$ and $B$ when represented
in terms of probability densities, thus enabling the identification of the more likely
model combination possibly explaining a new dataset.

Let $A$ and $B$ be two datasets described in terms of their respective probability
densities \emph{as well as} the associated support regions $\rho_A$
and $\rho_B$ as obtained for a chosen $\chi$.
The similarity between those two datasets can then be estimated as:
\begin{equation}
   \Lambda(A,B) = \frac{\int_{\rho_A \cap \rho_B} \, p(\vec{f}) \, d\vec{f}}
   { \int_{\rho_A \cup \rho_B} \, p (\vec{f}) \, d\vec{f}} 
\end{equation}

We again have that $0 \leq \Lambda() \leq 1$.    Observe that it is also possible
to assign normalized weights to each of the intersection regions between the support densities
in a new dataset corresponding to integration of the probability density within
that same region.  These weights can then be incorporated into the
above equation, so that the portions  of the support region of the new dataset that
explains a smaller fraction of the overall data population have smaller influence
on the decision.

It should be observed that the above described approach is still empirical, 
so that further formal validations should be developed.  It is also possible to consider
alternative methods for comparing between multidimensional distributions,
such as those involving adaptations of the non-parametric Kolmogorov-Smirnov
test (e.g.~\cite{MultKolm}).

The above obtained index provides a simple interesting manner for comparing a new dataset
$\omega_k$ with combinations of the existing datasets obtained through respective
set operations, in an analogous manner as done in the $M^*$ approach, but now also
incorporating the estimated probability densities respectively associated to each 
dataset.

Interestingly, the stochastic approach described in the present section paves the
way to other important capabilities, including the possibility to obtain not only 
a likely combination of models explaining a new dataset, but also the quantification
of how many elements of the latter relates to the each of the involved 
existing datasets.  This possibility, as well as the overall stochastic model
$M^{<\sigma>}$ described in this section, are further discussed and illustrated
respectively to a specific real-world dataset.  In  a sense, this type of more complete
description of a given dataset extends the concept of a dychotomic pattern recognition
decision to a relatively more complete model providing additional information about
how the new dataset explains and relates to the other existing datasets (and models).

\section{Case-Example: The Iris Dataset}\label{sec:case2}

We now address a typical case of supervised pattern recognition by using the
Iris dataset, which consists of 50 individual iris flowers from 3 species, each
being characterized by $N_f = 4$ features.  Thus, we have $N_{x} = 150$ individuals.
We shall be restricted to features 2 and 3 in order to allow the feature space to be more
easily visualized.

\begin{figure}[h!]  
\begin{center}
   \includegraphics[width=0.7\linewidth]{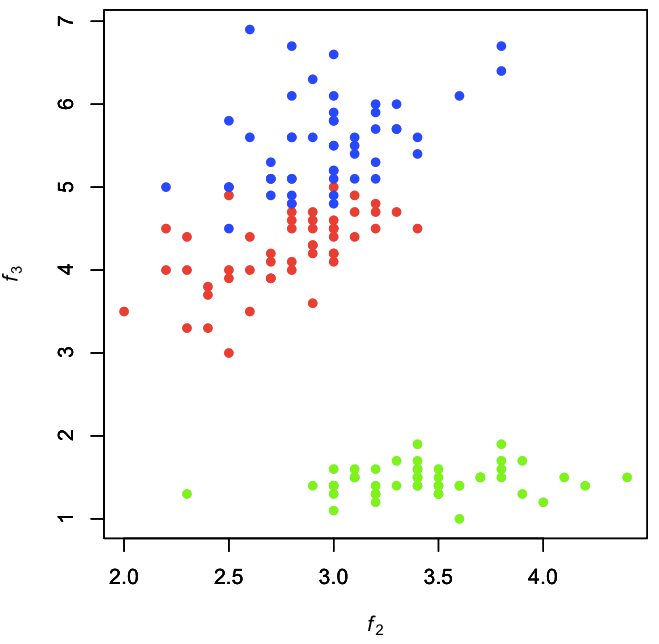}  
    \caption{The three species in the iris dataset shown in a features space
    derived from the original features $f_2$ and $f_3$: species 1 (green), species 2 (red), and 
    species 3 (blue). }
    \label{fig:iris}
    \end{center}
\end{figure}

Figure~\ref{fig:iris}(a) depicts the distribution of all the 150 individuals in the respective
two-dimensional feature space, with the three categories being identified by respective
colors.

The first step in our modeling approach consists of obtaining kernel expansions of the three groups of
points which, in the case of the present example, is achieved by using a circularly symmetric
gaussian as kernel assuming $\chi=0.97$.  The kernel expansion is then performed by convolving the original data
elements in each group, which are represented as Dirac's deltas, with the normalized gaussian kernel.
The adoption of a fixed percentile is reasonable given that the three datasets present a relatively
similar sparsity.  Observe also that, by varying the parameter $\chi$ multi-scale models of the
datasets can be derived.

\begin{figure}[h!]  
\begin{center}
   \includegraphics[width=1\linewidth]{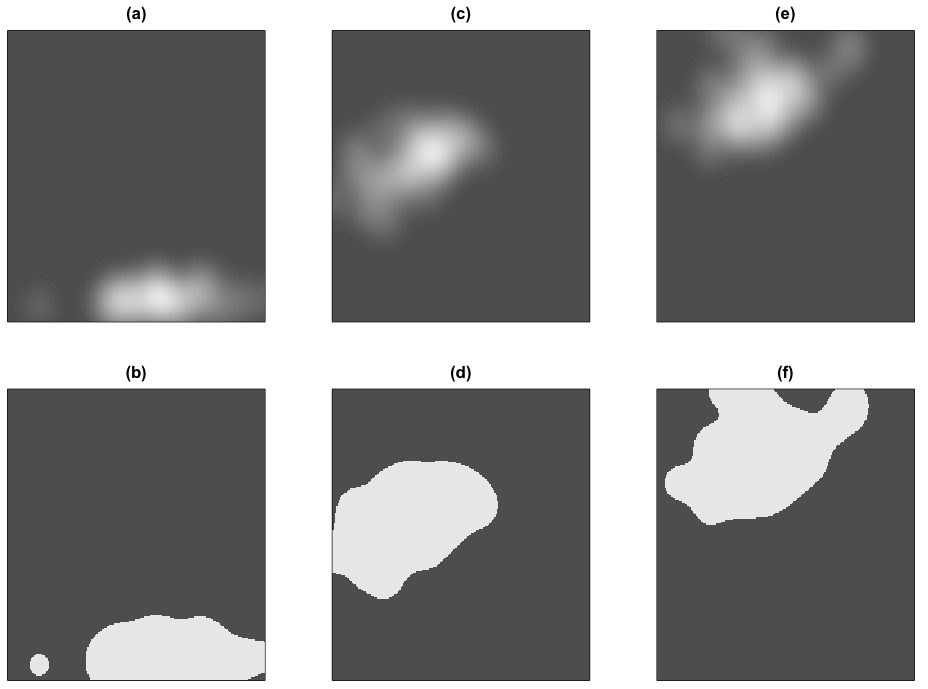}  
    \caption{The densities and support regions obtained by non-parametric gaussian
    kernel expansion of the three types of flowers in the iris dataset: $p_1(\vec{f})$ (a)
    and $\rho_i(\vec{f})$ (b); $p_2(\vec{f})$ (a) and $\rho_2(\vec{f})$; and $p_3(\vec{f})$ (a)
    and $\rho_3(\vec{f})$.  These results were obtained for $\chi=0.97$.  }
    \label{fig:kernel}
    \end{center}
\end{figure}

Figure~\ref{fig:kernel}(a) illustrates the result of the gaussian kernel expansion of each of the
three datasets in Figure~\ref{fig:iris}(b).

Figure~\ref{fig:kernel}(b) presents the three decision regions obtained considering  $\chi=0.97$.

The initial modeling framework is assumed to contain the models indicated in Table~\ref{tab:irispro}.

\begin{table*}[]
\renewcommand{\arraystretch}{1}
\begin{tabular}{|| c |c | c | c || c  |}  
\hline
& \emph{ textual description of the model}  & \emph{features/structures} & \emph{decision}  &  \emph{size} \\
\hline  \hline
$m_1$ & flowers belonging to species 1 &  $f_2$ and $f_3$  &  high similarity index for type 1 & 50 \\  \hline
$m_2$ & flowers belonging to species 2 &  $f_2$ and $f_3$  &  high similarity index for type 2 & 50 \\  \hline
$m_3$ & flowers belonging to species 3 &  $f_2$ and $f_3$  &  high similarity index for type 3 & 50 \\  \hline
\hline  
\end{tabular}
\renewcommand{\arraystretch}{1}
\caption{The initial modeling framework $M$ for the Iris case-example.}\label{tab:irispro}
\end{table*}

\begin{figure}[h!]  
\begin{center}
   \includegraphics[width=0.5\linewidth]{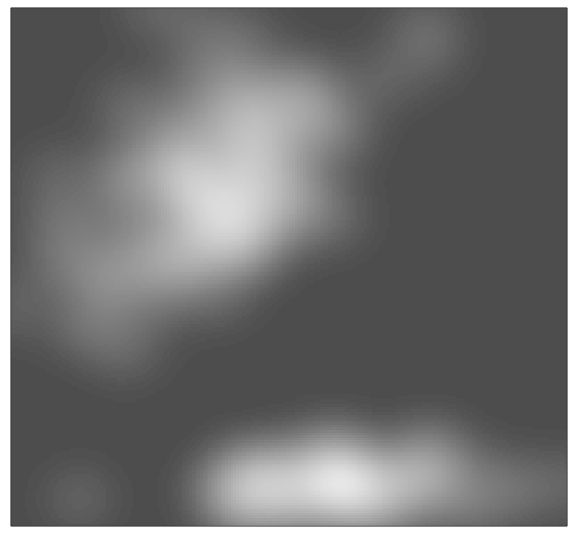}  
    \caption{The density probability function corresponding to the union of the three iris
    species.}
    \label{fig:iris3}
    \end{center}
\end{figure}

The density $p(\vec{f})$ corresponding to the union of the densities associataed with the
three iris species is shown in Figure~\ref{fig:iris3}.

\begin{figure}[h!]  
\begin{center}
   \includegraphics[width=0.8\linewidth]{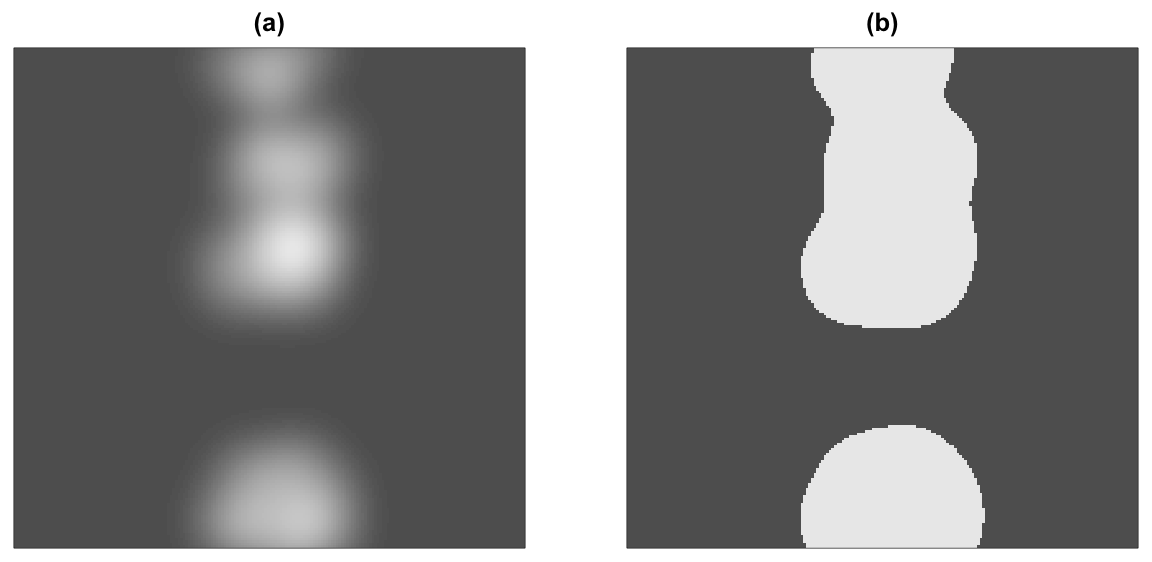}  
    \caption{The density (a)  and support region (b) of a new dataset. }
    \label{fig:data1}
    \end{center}
\end{figure}

Let's now assume that a new dataset $\omega_4$ becomes available, which is shown in 
Figure~\ref{fig:data1}.  

This set is then also kernel expanded by the same gaussian as before,   also using
 $\chi=0.97$.  The result is shown in Figure~\ref{fig:data1}(b).
Each new data element is then verified with respect to each of the three models, and updated
respectively.

It is now possible to perform a search for possible combinations of the existing models 
that best explain the new dataset $\omega_4$.   Among the several tried combinations,
up to the second hierarchical level of composition of logical conditions, the following dataset
was singled out as being the more likely to correspond to the new dataset $\omega_4$:

\vspace{0.5cm}
\fbox{\parbox{0.45\textwidth}{
\begin{equation}
   \omega_1  \; \cup \; \omega_2 \; \cup \; \omega_3  \nonumber \\
\end{equation}

\begin{itemize}
\item Overall similarity with $\omega_4$: $\Lambda =  0.375$
\setlength\itemsep{0em}
\item[ $\rightarrow$ ] 23  individuals related to  70.41 \% of dataset 1
\item[ $\rightarrow$ ] 19  individuals related to  41.88 \% of dataset 2
\item[ $\rightarrow$ ] 22  individuals related to  50.82 \% of dataset 3
\end{itemize}}}
\vspace{0.5cm}

\begin{figure}[h!]  
\begin{center}
   \includegraphics[width=1\linewidth]{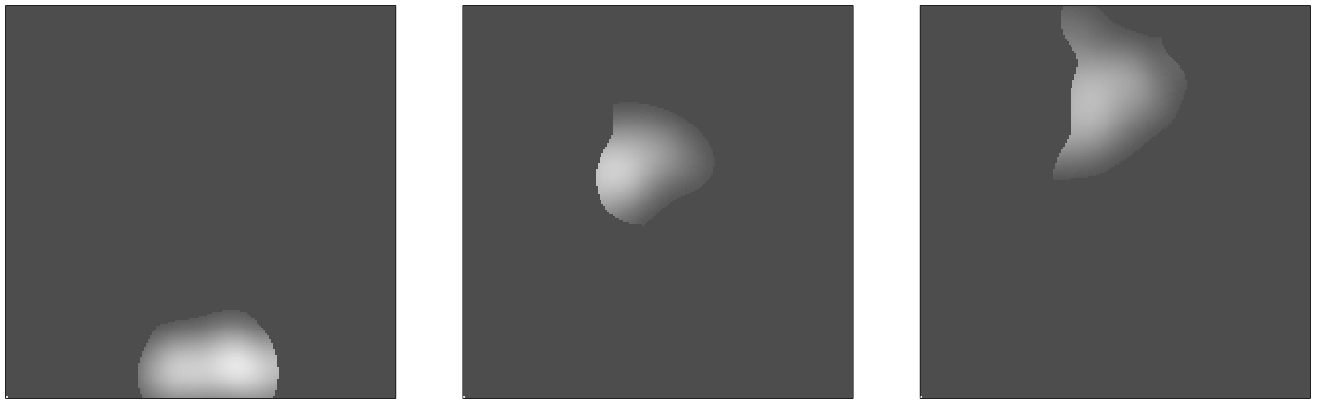}  
    \caption{The three probability density functions of the iris species clipped by
    the support region of the new dataset.}
    \label{fig:R1R2R3}
    \end{center}
\end{figure}

Figure~\ref{fig:R1R2R3} depicts the three density probability functions associated to
each of the three species after being clipped by the support region of the new dataset.
The integration of these clipped functions provides the identification of the 
relationship between the new dataset and the three existing models.

The number of individuals related to each of the existing
datasets correspond to the number of data elements contained in the
intersection between the support regions of the new dataset $\omega_4$ and
each of the other three existing datasets, which may to some extent overlap
one another.

In the light of these results, the new dataset cannot be considered to be
explainable by the union of the three original datasets corresponding to each of the
three iris flower species.  In addition, this new dataset seems to be more closely
related to the iris type 1, though relatively similar relationships are observed
also with the other two categories.  

As such, an alternative explanatory model in the domain of plant science
would need to be found or developed
for this new dataset.   In the case of this particular example, as the new dataset contains 
elements similarly related to each of the three original iris species, it could be conjectured
that the new samples correspond to physical alterations, such as a disease or changing
environmental or genetic conditions, taking place on the flowers and implying the feature 
$f_2$ to shift in a similar manner for all the three species.

Several other insights can be derived from the obtained descriptions as in the previous
example.  For instance, in case a new dataset is found not to relate directly to any of the
existing models while presenting a good relationship with the union of the respective
complements, it may be associated to the borders between the clusters in
a feature space.  Such interstitial regions can provide valuable information for identifying
effective separation regions in those spaces.  

\begin{figure}[h!]  
\begin{center}
   \includegraphics[width=0.8\linewidth]{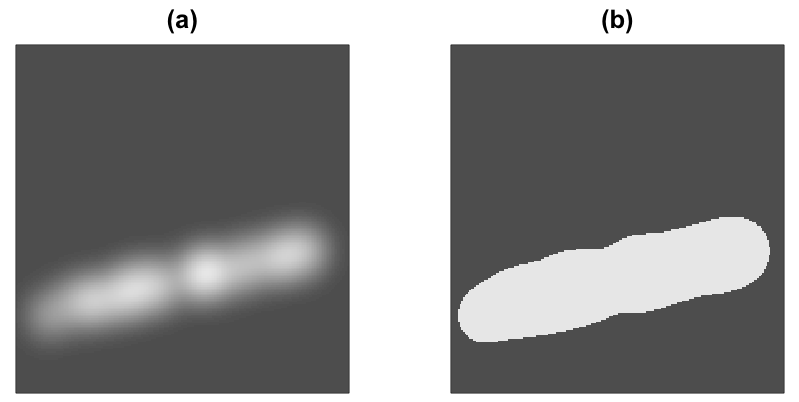}  
    \caption{The density (a)  and support region (b) of another new dataset. }
    \label{fig:data2}
    \end{center}
\end{figure}

As an example, consider a new dataset whose respective density and support region
is presented in  Figure~\ref{fig:data2}.

The application of the described approach yields:

\vspace{0.5cm}
\fbox{\parbox{0.45\textwidth}{
\begin{equation}
   \omega_1  \; \cup \; \omega_2 \; \cup \; \omega_3  \nonumber \\
\end{equation}

\begin{itemize}
\item Overall similarity with $\omega_4$: $\Lambda =  0.028$
\setlength\itemsep{0em}
\item[ $\rightarrow$ ] 0  individuals related to  0 \% of dataset 1
\item[ $\rightarrow$ ] 1  individuals related to  3.15 \% of dataset 2
\item[ $\rightarrow$ ] 0  individuals related to  0 \% of dataset 3
\end{itemize}}}
\vspace{0.5cm}

Therefore, the second new data can be understood not to correspond to the
model $\omega_1  \; \cup \; \omega_2 \; \cup \; \omega_3 $.  Unlike the
previous example, however, the minute number of elements related to any
of the existing datasets suggest that the second new dataset belongs to the
borders or interstices between the existing data.

\begin{figure}[h!]  
\begin{center}
   \includegraphics[width=0.5\linewidth]{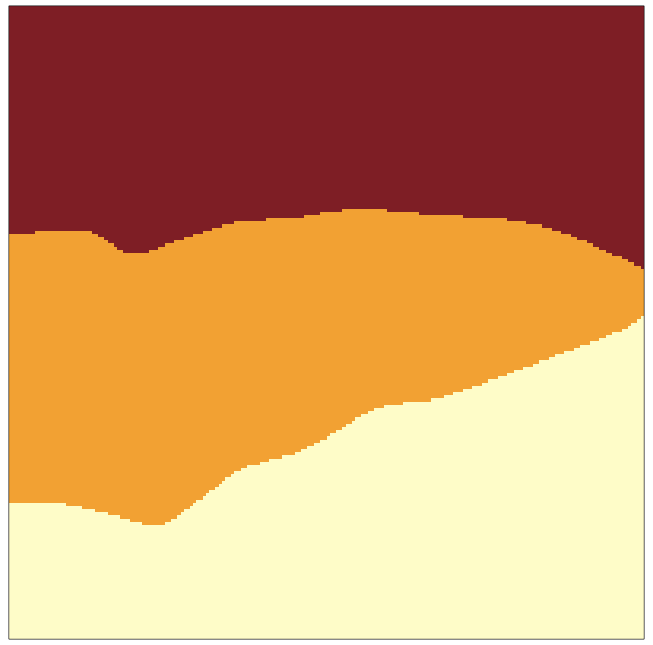}  
    \caption{The decision regions obtained by considering Bayesian decision theory
    with respect to the Iris example.  Observe the
    clipping of the overlap regions in shown in Fig.~\ref{fig:kernel}.}
    \label{fig:Bayesian}
    \end{center}
\end{figure}

In case the new dataset corresponds to a single data element, the similarity of the
$M^{<\sigma>}$ approach with Bayesian decision theory becomes more recognizable.
Figure~\ref{fig:Bayesian} illustrates the decision regions that are defined for the above
data.  

For each possible point $(f_2,f_3)$, the respective values are checked for every 
probability density associated to the existing datasets, and the label of the
existing dataset yielding the largest probability density value is associated to that
point.  This example considers equiprobable datasets, otherwise the mass density 
of the classes also would need to be taken into account (e.g.~\cite{shapebook,DudaHart}).  
Then, each new
data element can be classified as belonging to the dataset associated to the
label indicated by its respective features in the decision region.

Thus, for relatively narrow gaussian kernel expansion, the incorporation of new
data elements that precedes the checking for model combinations corresponds
very nearly to the classical Bayesian decision theory.    Though that approach
naturally integrates resources that can provide information about not only
data elements, but also datasets, as well as supplying information about the
adherence of each element with respect to the several existing dataset other
than the most likely one as well as the decision errors, it is felt that these
possibilities are not often realized, perhaps as a consequence of the focus
on dichotomic decision that is inherently motivated by the decision procedure.

\begin{figure}[h!]  
\begin{center}
   \includegraphics[width=0.5\linewidth]{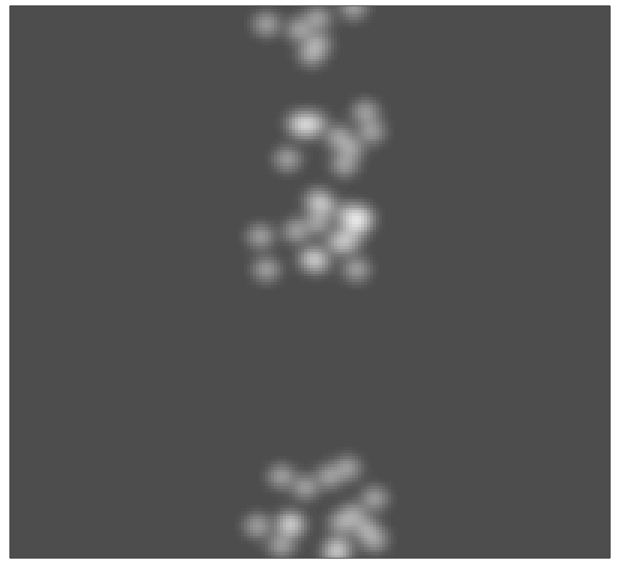}  
    \caption{The density obtained for the new dataset considering a much narrower
    gaussian kernel.  Observe that the density tends to converge to the original
    points (Dirac's delta) as the width of the gaussian is progressively reduced,
    also implying the $M{<\sigma>}$ approach to converge to the $M^*$ when the
    for infinitesimal gaussian width.  The non-infinitesimal width adopted in the
    stochastic case is necessary in order to allow non-zero probabilities in the
    probability densities describing each dataset. }
    \label{fig:iris_w3}
    \end{center}
\end{figure}

Interestingly, it can be shown  (see Figure~\ref{fig:iris_w3}) that the $M^{<\sigma>}$ 
meta model converges to the $M^*$ reference model as the kernels become infinitesimal.

It should be realized that while the analysis of datasets in 2D can be performed visually by humans,
the identification of most of the set combinations is typically difficult to be inferred in 
this manner, especially those involving combinations of set complements.  
The visualization of inference of set combinations in higher dimensional
feature spaces is even more challenging to be performed by human operators, therefore
providing even greater motivation for using automated methods such as the above developed.

Let's conclude this section by using the $M^{<\sigma>}$ approach to study the 
interrelationship between the three original iris dataset.  The results obtained for each
of these datasets are presented in the following:

\vspace{0.5cm}
\fbox{\parbox{0.45\textwidth}{
\begin{equation}
   \omega_1   \nonumber \\
\end{equation}

\begin{itemize}
\item Overall similarity with $\omega_1$: $\Lambda =  0.2735$
\setlength\itemsep{0em}
\item[ $\rightarrow$ ] 35  individuals related to  96.4914 \% of dataset 1
\item[ $\rightarrow$ ] 0  individuals related to  0 \% of dataset 2
\item[ $\rightarrow$ ] 0  individuals related to  0 \% of dataset 3
\end{itemize}}}
\vspace{0.5cm}

\vspace{0.5cm}
\fbox{\parbox{0.45\textwidth}{
\begin{equation}
  \omega_2   \nonumber \\
\end{equation}

\begin{itemize}
\item Overall similarity with $\omega_2$: $\Lambda =  0.4203$
\setlength\itemsep{0em}
\item[ $\rightarrow$ ] 0  individuals related to  0 \% of dataset 1
\item[ $\rightarrow$ ] 45  individuals related to  96.597 \% of dataset 2
\item[ $\rightarrow$ ] 16  individuals related to  41.393 \% of dataset 3
\end{itemize}}}
\vspace{0.5cm}

\vspace{0.5cm}
\fbox{\parbox{0.45\textwidth}{
\begin{equation}
    \omega_3  \nonumber \\
\end{equation}

\begin{itemize}
\item Overall similarity with $\omega_3$: $\Lambda =  0.4709$
\setlength\itemsep{0em}
\item[ $\rightarrow$ ] 0  individuals related to  0 \% of dataset 1
\item[ $\rightarrow$ ] 20  individuals related to  44.954 \% of dataset 2
\item[ $\rightarrow$ ] 42  individuals related to  96.477 \% of dataset 3
\end{itemize}}}
\vspace{0.5cm}

As expected, when taken separately, each of the regions resulted in relatively
low values of $\Lambda$.  At the same time, the well separated cluster defined
by region 1 has been corroborated by the fact that no relationship has been
found between this region and the others.   This is not the case with regions
2 and 3, which presented substantial mutual overlap.

\section{And Now, to the Features Layer}

It has been assumed in the $M^*$ approach that the features linking data elements and datasets
to respective models ensured a bijective mapping.  However, this is rarely observed
in many practical applications in which several features are to be taken into account.   

One point of critical importance here regards the fact that an association
between data and model being bijective depends on the context of the data.
This can be easily appreciated in terms of the following example.  Let's say that
we have a specific pair of glasses.  In case that object is used only inside one's
house (e.g.~for reading), there is no need to specify this pair of glasses very
completely, because there are no other similar objects to be distinguished from.
This is by no means the case in other situations, as when we are in a crowd, which
requires many more features to be specified.   In brief, the bijective mapping
of a dataset into a model also depends strongly on the respective environment $E$.

There is another equally important aspect to be taken while addressing features
in pattern recognition and modeling.  It has to do with the fact that some features
may or not be available or adopted.  In other words, two datasets that are actually
distinct may be decided to be the same because the distinguishing feature is not
available or even know.

In the light of these important concepts, we can now approach the problem of
features in the $M^*$ framework.   We have already seen that any entity, abstract
of real, needs to be first translated into quantitative or categorical measurements
before it can be recognized or associated to a model.  The data features has
been assigned a specific level in the suggested framework, namely the
\emph{features layer} $F$.

\begin{figure}[h!]  
\begin{center}
   \includegraphics[width=0.9\linewidth]{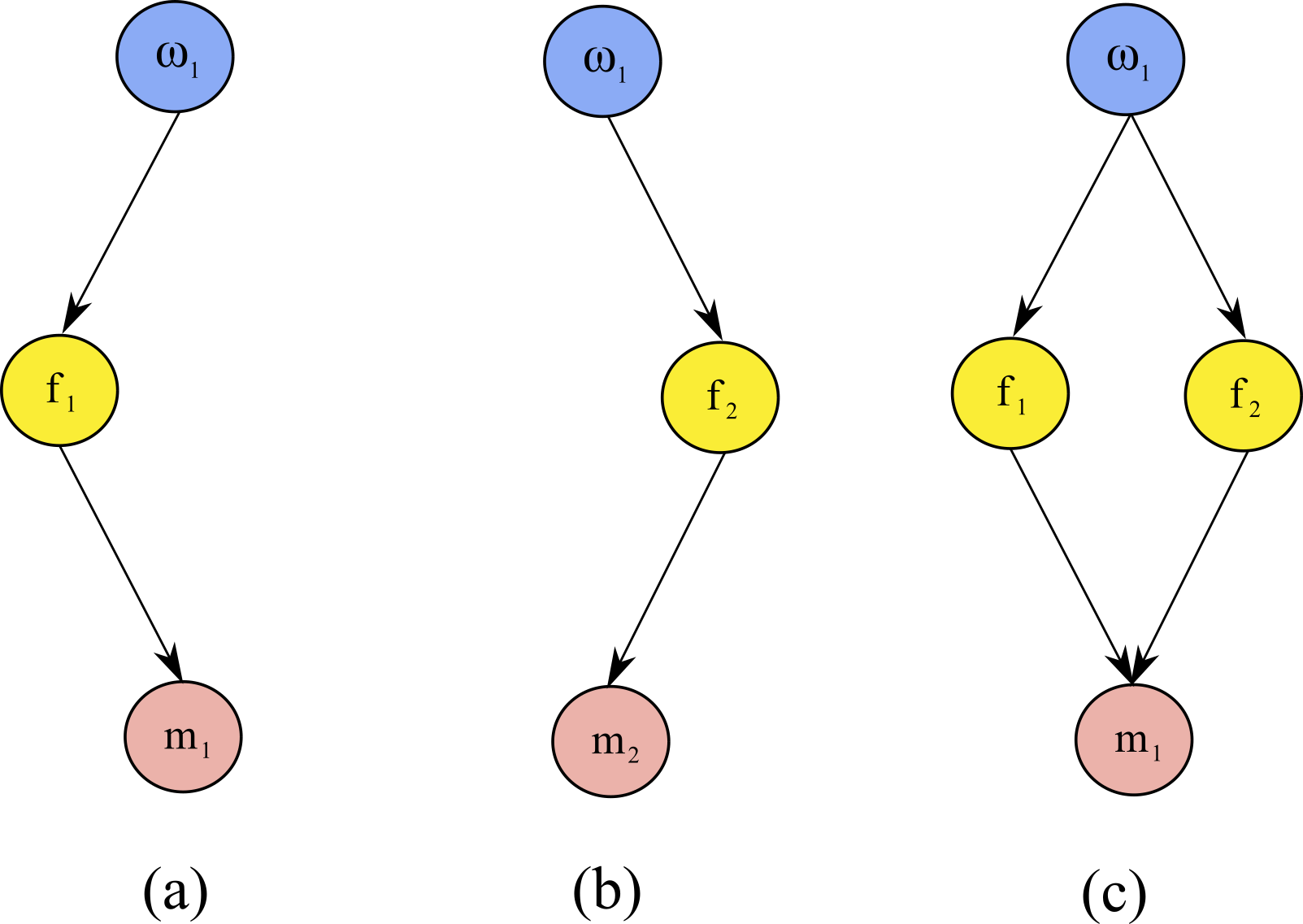}  
    \caption{A same dataset $\omega_1$ has been independently associated to
    a model $m_1$ when considering the feature $f_1$ (a), and to a model
    $m_2$ when taking into account the feature $f_2$. What can we say about
    the relationship between the models $m_1$ and $m_2$?}
    \label{fig:feat_conc}
    \end{center}
\end{figure}

Figure~\ref{fig:feat_conc} illustrates a situation that can substantially help to identify
and address how features can be treated in generalizations of the $M^*$ meta
model where the datasets do not necessarily relate bijectively with the respective models.

Let's assume that a dataset $\omega_1$ has been modeled into a respective
model $m_1$ while considering the feature $f_1$ (a), and also been associated to
another model $m_2$ while considering the feature $f_2$ (b).  What can be said
about the relationship between $m_1$ and $m_2$?

First, we have the situation in which both mappings are bijective, in the sense
that features $f_1$ and $f_2$ are each enough to specify the datasets.
In this case, the two features can be understood as being equivalent.
Recall that the environment $E$ influences the mappings being bijective or not.

Then, we have the case where all the data elements in $\omega_1$ share both
features, each of which establishing a non-bijective association between the
dataset and the two models.   In this case, the two models can be merged into 
the combined model shown in (c).  It is through these feature integrations
that a bijective mapping between a dataset and a model can be eventually
established.

However, in case only one of the data elements do now present
the same features, the two models should be treated as being distinct, being
therefore split into two subsets $\omega_{1,1}$ and $\omega_{1,2}$,
unless some tolerance is to be allowed. 

The co-existence of situations (a) and (b), which would therefore imply a non-injective mapping
from a dataset to two models, was ultimately resulted from overlooking of features
when the two models were developed independently.  Consequently, the consistency
of a modeling framework also depends on the careful consideration of the adopted
features. 

Another possibility in these situations is to disregard some features that are deemed not
to be of particular relevance to the model being developed.  Indeed, if taken 
systematically, the principle of splitting datasets would lead to not two real-world
objects being associated to a same model, for every two such entities will never
be completely identical.

The role of features in mediating the relationship between datasets and models can
be formalized in terms of set operations and propositional logic.  For instance:
\begin{equation}
   \omega_i \longleftrightarrow \left\{ f_1, f_2, f_3 \right\}  \longleftrightarrow m_i \nonumber
\end{equation}

states the fact that the dataset $\omega_i$ is bijectively related to the model $m_i$.

However, the situation:
\begin{equation}
   \omega_i \longrightarrow \left\{ f_1, f_2 \right\} \longrightarrow m_i \nonumber
\end{equation}

means that the joint consideration of the two features $f_1$ and $f_2$ maps in a non-injective
manner into the model, suggesting that these features are required but not sufficient for the
bijective verification of the model.

Algebraic logic combinations of features, also considering universal quantifiers over the data elements,
can therefore be found or better understood by taking into account such formal statements.  
For instance:
\begin{eqnarray}
   \left\{
   \begin{array}{l}
     \omega_i \longrightarrow \left\{ f_1, f_2 \right\} \longrightarrow m_i \\
     \omega_i \longrightarrow \left\{ f_3 \right\} \longrightarrow m_i  \\
     \forall \omega_j, j \neq i:    \omega_j  \longleftrightarrow \left\{ f_1,f_2,f_3 \right\} \nLeftrightarrow m_i \\ 
   \end{array}
   \right.
\end{eqnarray}

would imply:
\begin{equation}
  \omega_i \longleftrightarrow \left\{ f_1, f_2, f_3\right\}  \longleftrightarrow m_i \nonumber
\end{equation}

This result indicates that, given a dataset $\omega_i$ mapped into a model $m_i$ through a given set of
features, in case no other of the existing models has been bijectively associated with $\omega_i$,
then we can understand that the pairing $(\omega_i,m_i)$ satisfies a bijective association.

Another example, involving subsets:
\begin{equation}
   \left\{
   \begin{array}{l}
    \omega_{i,a} \subset \omega_i \Longleftrightarrow \left\{ f_1\right\} \Longleftrightarrow m_j \\
     \omega_{i,b} \subset \omega_i \Longleftrightarrow \left\{ f_2\right\} \Longleftrightarrow m_k  \\
    \omega_i = \omega_{i,a} \cup \omega_{i,b} \\
   \end{array}
   \right.
\end{equation}

implies that:
\begin{equation}
   \omega_i \Longleftrightarrow \left\{ \left\{ f_1\left[ \omega_{i,a} \right] \right\}, 
   \left\{ f_2 \left[ \omega_{i,b} \right] \right\}  \right\}\Longleftrightarrow  m_j \lor m_k \nonumber
\end{equation}

where $f_1\left[\omega_{i,a} \right]$ means that the feature $f_1$ is restricted to the subset $\omega_{i,a}$.

It is also possible to extend this logical formalization to the data elements level, allowing
the consideration of features related to specific subsets of $\omega$.




\section{And How About Clustering?}

It has already been observed in this work that there are two types of pattern
recognition: supervised and unsupervised.   As only the former has been
considered so far in our approach, additional considerations can now be
developed regarding the also important subject of
unsupervised classification, which is also typically known
as \emph{clustering} (e.g.~\cite{DudaHart,Koutrombas,Luxburg,Hennig,CostaGener,evaluate_network}).

\begin{figure}[h!]  
\begin{center}
   \includegraphics[width=0.9\linewidth]{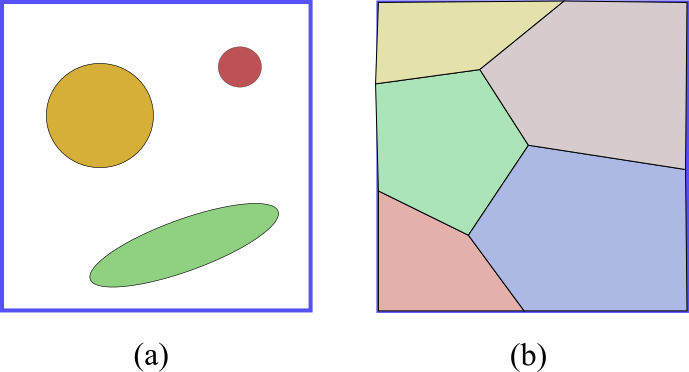}  
    \caption{An example of clustered data  containing 3 groups of points
    characterized by being well-delimitated and separated (a), as well 
    as another type of data separation which,
    though not having interstitial regions, still ensures all the data elements to be
    properly compartmentalized and classified (b).}
    \label{fig:clustering}
    \end{center}
\end{figure}

Generally speaking, clustering consists  basically of finding separations our groupings
between the respective distributions of data elements in the adopted feature space, 
as illustrated in  Figure~\ref{fig:clustering}(a).  

In another related situation, also depicted as the situation (b) in Figure~\ref{fig:clustering}, 
the data elements are perfectly compartmentalized into respective categories or models, even
though there are not interstices.   The groups are also adjacent one another.
This type of situation is rarely considered in 
clustering, because of being impossible to solve while considering only the spatial
distribution of the data elements.     As a matter of fact, the identification of clusters
in cases where the datasets of interest are not well separated represents a substantial
challenge in pattern recognition, because of the difficulties implied.  At the same time, 
several real datasets tend to present overlaps and 
adjacencies, as is the case with the iris dataset in Figure~\ref{fig:iris}.

Figure~\ref{fig:clustering}(b) depicts an interesting situation that helps to understand how
the fact that a subset satisfies a model does not necessarily imply or relate to a cluster.  

\begin{figure}[h!]  
\begin{center}
   \includegraphics[width=0.4\linewidth]{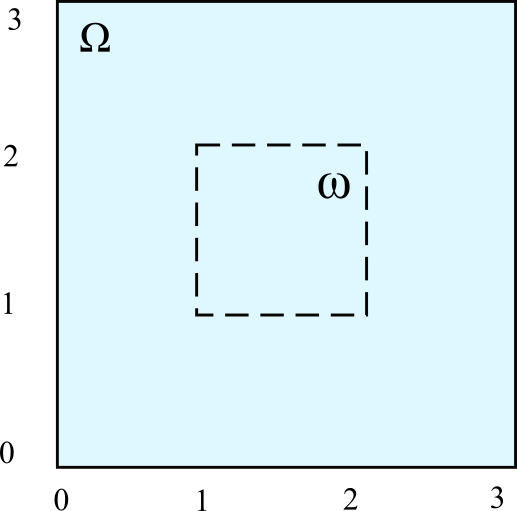}  \hspace{0.3cm}
   \includegraphics[width=0.4\linewidth]{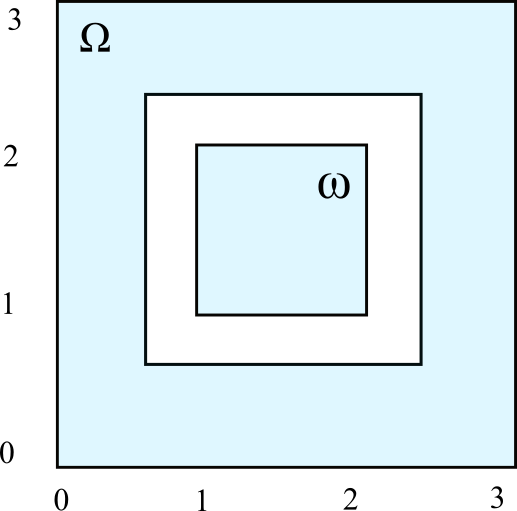}  \\
   (a) \hspace{3.5cm} (b) 
    \caption{An example (a) of dataset $\omega$ which, though perfectly explained by 
    the model  $m: (x,y) | (x\geq 1) \land (x \leq 2) \land (y \geq 1) \land (y\leq 2)$, is not a cluster.
    In (b) we have a dataset $\omega$ that both is explained by this same model, but which is
    also a cluster.  There is not need that a dataset corresponding to a model be a cluster,
    though clusters may motivate a dataset to be explained.}
    \label{fig:noclust}
    \end{center}
\end{figure}

Here, $\Omega$ corresponds to a portion of  $\R^2$, so that the possible data elements
are ordered pairs $(x,y)$, with $x,y \in \R$.  The whole dataset $\omega$ delimitated in the
figure satisfies a well-defined model, namely $m: (x,y) | (x\geq 1) \land (x \leq 2) \land
(y \geq 1) \land (y\leq 2)$.  None of the points outside $\omega$ belongs to this model,
so that we also have a perfect \emph{partition} of $\Omega$ defined in terms of $\omega$
and $[\omega]^C$.  Yet, despite all these specific and distinct properties of $\omega$, it
is in absolutely no way clustered, or present interstitial regions with the remainder of the
points in $\Omega$.   The main effect of the interstice is to call our attention on $\omega$,
motivating the explanation in terms of model that was already valid.

The above examples helps us to realize that both the datasets that are well-separated as well as
the other types of compartmentalized datasets can be explained by models.  In this sense, a 
dataset being a cluster
is a property that would be independent of having a model associated to it or not.

A particularly interesting relationship between clustering and modeling relates to the fact
that clustering provides one of the main means through which delimitated datasets can
acquire enough interest as to motivate respective explanation through modeling.  The
association of  specific properties of interest (e.g.~a well-separated
group of bacteria capable of digesting some specific material) to a clustered dataset 
tends to make it even more likely to be modeled.  It should be also observed that such
distinguishing properties are primary candidates to be adopted as part of the
feature for the specific characterization and modeling.

Another possibility worth considering is that, once a not necessarily well-separated
dataset is identified as having special importance, it may become a more isolated group
as a consequence of identification of more discriminative features, or even as a consequence
of actions motivated by the need to separate the data, such as performing features
transformations.  For instance, a plant species that is initially little
different from others, but present some interesting property, may be selectively breed
to the point of being transformed into a more separated cluster.  It is also possible to contemplate
the situation in which actions are taken for reducing the cluster separation.   

Yet another possible mechanism leading to the creation of clusters is as follows.
A single, or a few, data elements are observed to present a given property of interest.
Efforts are then invested in identifying more elements satisfying that property, but
without consideration of a control counterpart.  As a consequence of this biased
procedure, new data elements will be identified that present the desired property
which will, as a consequence, yield a well-separated cluster, because the possible
features that would possibly imply be adjacent in the selected feature having
been filtered out.

\section{Malleability of Datasets and Models}

Given a structure represented as graph that is subjected to topological changes
such as inclusion/removal of edges or nodes, it is possible to estimate the
potential of this graph to undergo distinct successive changes.  This can be
done by using the recently introduced \emph{malleability} measurement~\cite{Malleability}. 

Because the $M^*$ framework can ben represented as a graph, it becomes
interesting to characterize and compare the potential of distinct modeling frameworks in 
terms of their respective malleability.  

The main problem when devising means to quantify the malleability of a graph or
network concerns the fact that two or more of these structures (e.g.~two
separated instances of a network along time), as identified in
terms of labels associated to the respective nodes, may actually correspond to
the same topological structure, differing only with respect to the associated labelings,
a property known as \emph{isomorphism},

As it is often very computationally expensive to be decide whether two or more
graphs are isomorphic, a viable alternative is to compare those networks after
they have been mapped into a set of features.  Remarkably, this mapping does
not need to be bijective, provided we remain limited to comparing the networks
from the perspective of the adopted features, which underlies the approach
reported in~\cite{Malleability}.

Let $\gamma$ be a graph, and let a specific manner to change this network
be chosen.   At a given time instant $t$, after the application of every possible
instance of the considered change (e.g.~removal of any of the possible edges),
a total of $D$ distinct networks are found to be derived from the initial configuration $\gamma$,
each with a specific probability $p_i$.   The malleability of this network can be
calculated as:
\begin{equation}
   \mathcal{M_{\gamma}} = e^{-\sum_{i=1}^{D} p_i log_2(p_i)} = e^{\eta}
\end{equation}

where $\eta$ is the entropy of the probabilities $p_i$.

It is posited here that this index provides a good way to quantify the potential
of a modeling framework to be adapted for inclusion of new datasets and/or models.
At the same time, it also supplies an objective means for characterizing the
adaptivity and robustness of a given modeling framework.

\section{Complexity}

\emph{Complexity} 
(e.g.~\cite{Waldrop:1993,Kauffman:1996,Heylighen:1996,Lofgren:2007,Edmonds:1995,Immerman:2015}) has remained a great 
challenge to be defined in an 
ample and yet accurate manner.    This is all the most remarkable given
the great importance this concept has achieved not only
in scientific and technological fields, but virtually in all human activities.

Though continuing efforts have been made at grasping what
complexity means,  many of these have been conceived in order to 
address relatively specific problems by using respective concepts and approaches.  
A review of some of the main approaches to quantifying complexity can be
found in~\cite{CostaCompl}

More recently, an attempt has
been made at obtaining a more comprehensive and flexible definition of complexity
that would remain compatible with the way it is more generally
understood by humans~\cite{CostaCompl}.  The underlying idea is to relate complexity 
to the \emph{costs} of developing and operating/maintaining a model.
Given that the concept of cost was conceived precisely to adapt to
relative variations of specific resources availability and demands along
time and space, the cost of a model seems to provide a
particularly interesting perspective from which to approach the complexity
involved in modeling.

We have already seem that several events and facts conspire to limit
the modeling approach, such as described in the models reported here
(Section~\ref{sec:errors}).  Some of the most relevant of those are now
briefly discussed as indicators of complexity.

First, we have that real-world $\Omega$ universe sets tend to be extremely
large, as it is characteristic even for specific types of plants and animals.  
The obtention and maintenance of these large datasets imply not only
computational expenses, but also curation by experts.  

The fact that the total number of subsets derivable from $\Omega$ corresponds to
$2^{N_{\Omega}}$, where $N$ is the number of elements in $\Omega$, a
combinatorial explosion soon takes place that makes unfeasible to consider systematic
modeling approaches taking into account a substantial portion of the total possible
number of datasets.  Therefore, even in cases where the individuals are enumerable,
and in absence of sampling and other types of error and noise, it becomes necessary
to resource to optimization techniques capable of selecting particularly interesting
subsets out of an extremely large number of possibilities.  This implies substantial
development and computational costs, and it is poised to result in local minima,
all of which contributes to making the modeling expenses considerable, accounting
probably to many sources of complexity typically associated with modeling.

Then, we have situations in which the original data elements are too similar one another,
implying a large number of features to be derived, some of which will probably imply
in relatively high experimental costs.   A related problem implying expenses and complexity
regards the situations in which some of the individuals in $E$ or $\Omega$ are rarely
found.

To the above can be incorporated other several types of errors, noise, sampling and
other limitations discussed in Section~\ref{sec:errors}.  We can therefore conclude that 
modeling at a more extensive and accurate level
can becom extremely expensive and complex in several ways and situations.
As discussed in Section~\ref{sec:errors}, creativity could be one of the best antidotes
to complexity, allowing interesting results to be obtained even in challenging situations.

\section{Complex Networks} \label{sec:netsci}

With a history going back to the beginnings of humanity (e.g.~maps), passing
through the K\"onigsberg bridges, graph theory, and sociological research, the
subjects covered in the area of network science (e.g.~\cite{netwsci,networks:2010}), 
which focuses on complex networks, took off with
studies of the Internet and the WWW.  Briefly speaking,
the subject of study of this area concerns graphs that present a topology that
cannot be described in terms of one or few topological measurements such as
the node degree (e.g.~\cite{Costa:CDT2}).  Therefore, a complex network would tend
to present topological features markedly distinct from a regular graph or a stochastic
counterpart such as a uniformly random network.

The remarkable success of this area, not only from the applied but also theoretical
perspectives, resides greatly on the ability of graphs to represent virtually every discrete
structure or phenomenon, also allowing for the fact that even continuous structures can
be discretized to some resolution level.  Another welcomed aspect highlighted by
network science consists in its motivation for studies integrating the topology and
dynamics of complex systems.

Given the above mentioned features of network science, it becomes particularly
interesting to discuss, even if briefly, the relationship between complex networks
and the meta modeling framework suggested in the present work.   This is
the subject of he present section which, however, shall focus on the issue that
complex networks obtained from real or abstract datasets have been frequently
derived while considering the \emph{similarity} between the properties of the
data to be explained (e.g.~\cite{SimilarityPattern}).  For instance, in an
informational network where the nodes stand for specific documents such as books,
web pages, or works of art, the interconnection between these nodes is often
performed while taking into account the content similarity, or overlap.  Other
examples involve networks constructed while considering the similarity between
the features characterizing specific entities associated to nodes, such as in
networks of living species, stars, shapes, etc.  The similarity, or distance, between
pairs of node is often represented in terms of weights associated to the respective
edge.

A first important point here regards the fact that the very act of associating a node
to an entity to be represented actually corresponds to identifying a model for
that entity, which needs to be done in terms of a set of features.  Then, these
features can be compared, typically through similarity, while interconnecting the nodes.

While these approaches have great interest and potential, as already
been demonstrated by the large number of well-succeeded applications
(e.g.~\cite{surv_appl}), the often considered networks correspond only to 
one possible manner in which entities can be related, more specifically through
similarities or distances.  The framework describe in this work seems to allow
several possibilities for extending the network-based representations.
This corresponds, basically, to employing the several combinations of models
obtained while deriving a network, involving several types of set operations and
logical constructions.  For instance, it is possible to connect two nodes corresponding to
respective datasets with the node associated to the union, difference, etc., between
these two datasets.  Each of these connections could be identified by a respective
label corresponding to the respectively applied set operation.

The integration between network science and the $M^*$ framework can proceed
mainly by considering both the individual data elements and respective datasets
through the respective combinations allowed by the exact or approximated
combination of datasets and models.  In approximated cases, it may be of particular
interest to compare networks representing an exact modeling framework with
the available datasets that can be approximately explained by each of the theoretical
models.  Other possibilities already hinted in this work are to derive bipartite
networks from the association between data elements and respective models, 
as well as considering the hierarchical constructs resulting from combinations of
data or models as networks, to which similarity links may also be added.

Though the possibilities are many to be identified and discussed here, it is
hoped that the above discussion may motivate further related analysis and
developments.

\section{Collaborative Research}  \label{sec:collab}

The $M^*$ meta model provides several interesting subsidies that can be employed to
obtain insights about the characteristics and challenges in collaborative research.
Of particular importance here is the requirement in the $M^*$ metal model of 
keeping full compatibility and consistence between datasets and models.

Science has largely relied on the integration of two complementary approaches:
individual and collective.  In the former case, we have
a single scientist, possibly with the assistance of a team, working on specific problems.  
The latter approach is characterized
by more ample collaborative initiatives involving big projects, regular meetings, and, more recently, 
WWW-based resources.

While the case of individual research can be directly related to the
the meta models suggested in the present work, the collaborative counterpart
requires more analyses.  In fact, it should be observed that a fully individual
research initiative is virtually impossible, as one needs to learn and to communicate
concepts and results.

\begin{figure}[h!]  
\begin{center}
   \includegraphics[width=0.8\linewidth]{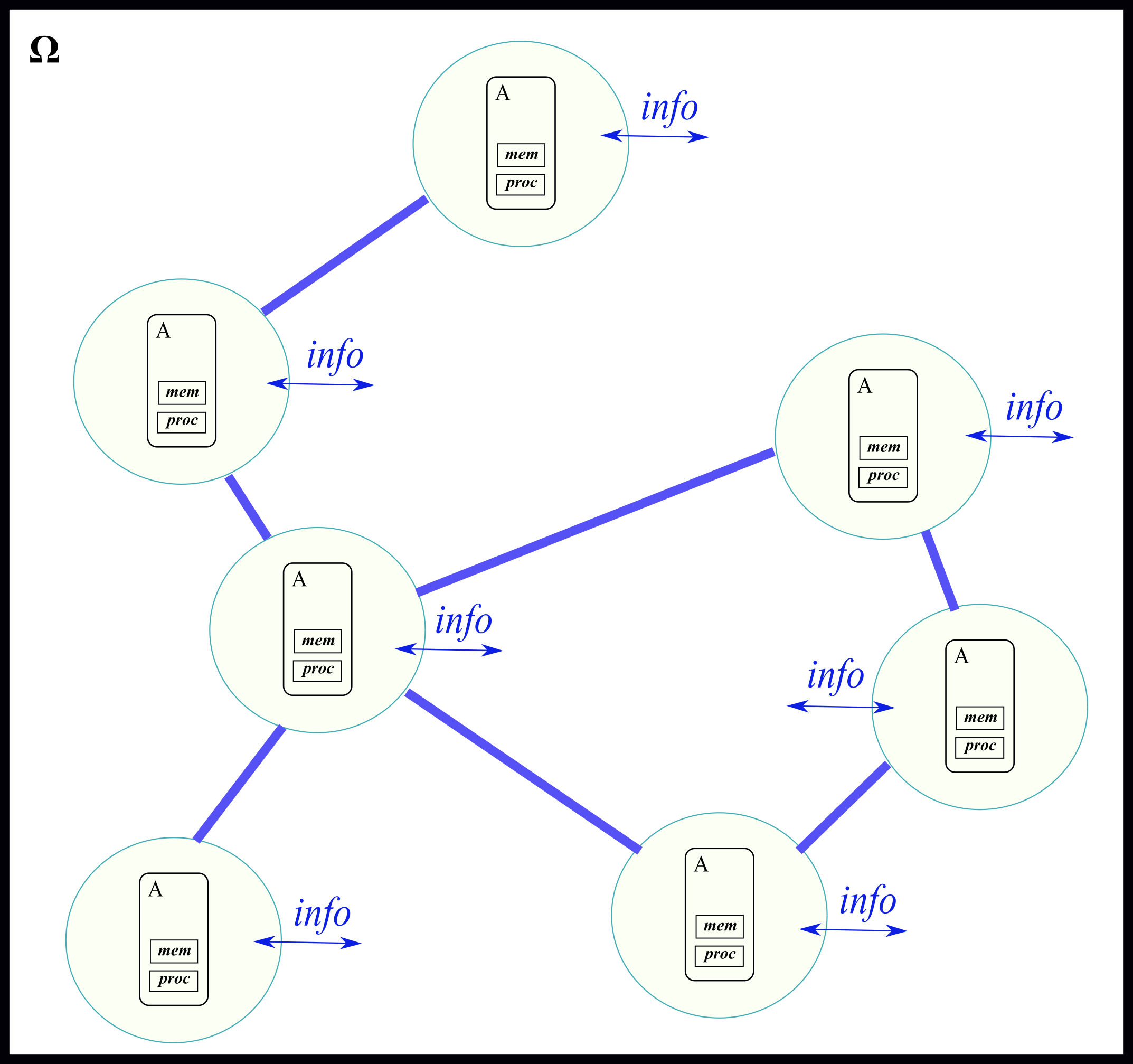}  
    \caption{A highly abstracted and simplified model of interaction between
    agents (living beings or machines) that develop models. Issues of particular relevance 
    for collaborative research regards how well integrated the agents are, which depends
    on the network topology, as well as the importance of keeping consistency between
    communication, datasets and models not only at the individual agent level, but also among all the
    modelers.  Two possible ways to implement the latter include through shared 
    external resources as dabatases, or by implementing continuing exchanges of
    information between the agents.}
    \label{fig:collab}
    \end{center}
\end{figure}

Figure~\ref{fig:collab} illustrates a highly simplified situation involving several
agents like that in Figure~\ref{fig:individual} that can collaborate one another through the
represented network of information exchange.

As each researcher gathers new data and develop respective
models, it becomes important to \emph{communicate} these findings as wide as
possible through the network, so that the results can be further validated and
the other modeling frameworks can be
updated and kept consistent.    This requires a \emph{shared},
\emph{standardized} or \emph{translatable} (e.g.~via meta models) body of
datasets and models.  At the same time, it is essential that
the same features are validated and adopted by all or most researchers.
The larger the modeling framework, more possibilities can be tried while
combining and integrating the respective models, though adding substantially
to the overall complexity.

Of particular interest becomes the possibility to construct shared \emph{databases}
integrating all existing datasets and respective models, as well as employing
\emph{automated} means for implementing or assisting the modeling activity.   The latter is
of critical interest because the typical amount of information and knowledge
currently necessary for many areas by far exceeds the cognitive and memory
capabilities of any human being.  Special  attention should be also given 
to establishing common data and modeling formats and representations,
especially concerning the identification of suitable data structures to be adopted
in each specific situation or generalized as shared resources.

Other challenges in collaborative research involves how to cope with the
several types of modeling errors discussed in this work, to which can be
added errors and limitations of the communication through the existing
network.  Data and model \emph{curation}, possibly assisted by automated
means, could contribute to 
achieving reasonable levels of data and model quality.  As a matter of fact, 
it is also important to continuously keep and expand the communicating 
resources.

\section{Deep Learning}

Similarly to network science (e.g.~\cite{netwsci,networks:2010,Costa:CDT2}), deep learning (e.g.~\cite{CostaDeep}) 
has achieved substantial
acceptance and success in a relatively short period of time, while also relying
on approaches going back to the 19th century, a great deal of which related to the
neuronal network paradigm.

The success of deep learning stems mainly from the fact that it paved the way to 
solving many problems that had remained as big challenges for pattern recognition.
This has been achieved thanks to several factors (e.g.~\cite{CostaDeep}), including the
consideration of vast amounts of data and computing resources, as well as the
development of new and creative concepts and methods.  Most deep learning
systems have the neuronal elements arranged in a several sequential and/or
parallel layers containing a vast number of components.

A typical deep learning system can be understood to involve a vast number of basic
processing elements, analogous to neurons (e.g.~\cite{CostaNeurPatt,Haykin}), 
that are successively applied
(often through convolution) onto the incoming data in order to derive valuable features 
and perform successful pattern recognition.  

Following the approach reported in this work, it should have become evident that the 
development of formal $M^*$ models requires strict maintenance of dataset
and model consistency not only within themselves, but also one with the other,
while also involving the ample consideration of every data element in the environment
$E$ at all times.   For proper operation of the described modeling approach, 
every data element and model would need to be considered and related along the whole
activity of modeling.  This fact may well be related to the critical importance of 
taking very large training dataset and processing resources as it is
characteristically found in deep learning.

Though remarkably successful in many applications, deep learning also has its
respective challenges, including the relative difficulty oin inferring the rules through which the
solutions have been obtained, or translating the obtained trained parameters into
formulations that can be more easily communicated to humans.   

Given that the $M^*$ and other related approaches described in the present work
provide a relatively formal description of how datasets can be mapped into models,
they may pave the way for identifying related methods for inferring the learned 
classification rules and translating them into more tangible statements.   This could
be done, for instance, by trying to associate formal or textual models to some of the datasets 
that have been assigned to categories by a respective deep learning system and then
trying to explain other, more complex datasets, in terms of logical combinations between
the identified datasets.   The $M^{<\sigma>}$ stochastic variation of the $M^*$ approach
may be of particular related interest, as it is directly related to the decision regions
normally associated to the basic deep learning processing elements.  

It may also possible to conceive manners of integrating the $M^*$ framework within a
deep learning system, so that the recognition of the input datasets can be directly
accompanied by the identification of possible respective models.
Another interesting possibility would be to try to adapt the impressive hardware resources 
developed for gaming and deep learning to perform the basic $M^*$ operations and 
manipulations in a faster manner.

\section{Creativity}

As complexity and other words that have received great attention, \emph{creativity}
has also proven to be a challenge to being defined and characterized.  Here, we
will adopt the approach described in~\cite{CostaCreatCompl,CostaCreat}, more specifically that creativity corresponds
to manners of achieving effective results with relatively great efficiency, little cost, and great innovation.  
It is from this perspective that we will briefly discuss how modeling, especially as approached
in the $M^*$ initiative, is related to creativity.

One of the main ways in which creativity can be achieved consists in seeking for analogies
or metaphors between two or more problems that, though belonging to different areas, present some
interesting similarities and analogies.  An extremely simple illustration of this type of creative 
association is the pairing of
numeric sequences such as $1, 2, 3, \ldots,$ with successive letters as $a, b, c, \ldots$

A particularly interesting point is that the $M^*$ modeling framework is critically dependent
on establishing an effective bridge, through a strict bijective association,
between set operations in the dataset domain and logical 
manipulations in the modeling domain.   In addition, by providing possible logical and
mathematical explanations that can be eventually translated into textual statements
defining a model, the proposed framework contributes to making those dataset more
accessible and tangible to those interested in their respective analysis and modeling.

In addition, engines analogous to those illustrated in this work may be employed as
means of providing insights about possible relationships between models and 
datasets, as well as clues on how to combine or develop new models capable of
explaining new datasets.

Following similar reasonings, the consideration of concepts and methods from a large
number of areas adopted in the development of the $M^*$ framework also contributes
to identifying possible creative analogies between their respective properties, challenges
and advantages.

Additional aspects of the $M^*$ approach that may favor creativity include the 
establishment of relationships and similarities between the several datasets, which may
be related to different fields.  Thus, the basic set and logical operations underlying the
$M^*$ operation can be understood to be directly related to potential creative associations
between different datasets, areas, features, and types of models.  The fact that several
types of features and models may be incorporated into the suggested meta models also
contribute to provide grounds for creative investigation.

Another not so directly identifiable aspect concerns the fact that the $M^*$ approach 
evidences the breathtaking combinatorial complexity involved in model building for relatively 
large, or even moderate, $\Omega$ sizes.  This huge complexity can also understood as
contributing more space and degrees of freedom while seeking for creative approaches and
solutions.  Interestingly, complexity and creativity seems to be in a sense intrinsically connected
and interdependent, even though they often opposes one another.

\section{Concluding Remarks}

And so we have reached the conclusion of the present work.  It has been a relatively
long development, as implied by its wide 
main objectives of taking into account many concepts and areas 
as the main subsidy for developing a putative meta modeling approach that could
provide some insights about  model building, decision taking,
and pattern recognition, among other possibilities.

We started by discussing what we called the \emph{informational schism} that is unavoidably
established between the real world and any modeling agent, be it a living being or a machine.
As it has been argued, the appearance of these agents was only have been allowed by the 
creative incorporation
of modeling abilities capable of providing effective means for interacting with the respective
environment.   This modeling ability is particularly critical because it ultimately provides the
means for taking effective decisions on subsequent actions based on previous experiences
and the consideration of current environmental conditions.

Because of the central role of model building in so many areas, including pattern recognition,
it becomes interesting to develop respective
abstract models capable of providing some insights and resources for modeling. 
The main requirements that such a meta model should have were then identified and
listed, including many constraints and sought properties.

The critically important task of mapping data into models was then approached, with
special attention being focused on the need to preserve as much information as possible,
which was shown to be only fully possible in case an bijective association is established
between the existing datasets and the respective models.  This relationships can be
achieved by always ensuring that every element in a dataset is satisfied by the respective
model, and vice-versa, therefore implying in a respective bijective mapping.  It has also
been shown that the current dataset environment, as well as the choice of features,
also contribute in defining whether a given
association between dataset and model is bijective or not.

At the same time, the generalization provided by each model in explaining all the elements in the
associated dataset, which requires a non-injective mapping, was accommodated in the
fact that this non-bijective relationship is maintained at the level of
$(dataset,model)$ associations.  The incorporation of parameters and features, two
important components of model building and pattern recognition, was also discussed
and addressed.

Subsequently, we identified and briefly considered some of the main sources of limitations in
achieving a fully complete and precise model, which include the presence of
noise, sampling, and errors.
The characteristics and effects of these limitations were then discussed respectively to the
main components and actions involved in model building.

Having thus obtained subsidies from the  presented and discussed several 
points involved in modeling, decision making and pattern recognition, we 
started an approach to developing a relatively formal meta modeling framework, which was 
called $M^*$.  

This framework provides several interesting features that emanate from 
the strict bijective association established between data and models, including the
definition of a bridge between these two worlds and the derivation of a paired algebra
of datasets and models which can be employed to find models for new datasets
through the logical combination of model statements as well as in terms
of set operations between the existing datasets.   The 
$M^*$ approach was then illustrated respectively to a case-example involving datasets
composed of patterns derived from binary lattices.  In fact, despite its seemingly
strict requirements, there are several problems characterized by relatively small
amounts of discrete data that can be approached by using the $M^*$ framework.

The requirements underlying the $M^*$ approach were then progressively relaxed 
as the modeling approach was extended to cope with some of the severe limitations that 
are characteristic of pattern recognition and scientific modeling, including noise, errors
and sampling.  This led to the $M^{<\epsilon>}$ and $M^{<\sigma>}$ meta models, 
the former accounting for comparing data in presence of error or sampling, while the latter
incorporating  means for dealing with frequently necessary stochastic description of datasets 
in terms of probability densities.  Both these models were then illustrated respectively to a 
elementary number theory and a real world data set related to the iris flowers.

Though we have seen that pattern recognition corresponds to a kind of modeling, it
often has the specific characteristic in which the condition to be satisfied
corresponds to the pertinence of the respective dataset with a given category.
Oftentimes, but not always, the result of pattern recognition does not incorporate
a more complete description of why the respective dataset has been decided to
be assigned to a certain class.  In addition, the data environment $E$ tends to
be constrained in practical pattern recognition problems, while in scientific modeling
$E$ is usually assumed to extend as much as possible toward the whole physical world.
Nevertheless, pattern recognition remains an instantiation of model building.

In addition to providing insights about the intricacies of modeling, the suggested
frameworks may also be used to derive practical methods and software engines
for automated or assisted scientific modeling and pattern recognition, among 
other possibilities.

At the same time, the presented developments also emphasize the need to keep
data and models as much as possible consistent and integrated, which poises some
specific challenges regarding formats, data integrity, validation, among other issues.

The important related subjects of clustering, complexity, collaborative research, deep
learning, and creativity were then considered and discussed in terms of several of the
concepts and insights provided by the reported modeling framework.

The many implications and possibilities allowed by the presented
concepts and methods pave the way to a substantial number of possible future
developments.  These include, but are not limited to, further extending the
family of models derived from the reference $M^*$ framework so as to be able to
address additional constrains, developing effective concepts and methods that can
be used for implementing kernel expansion in higher dimensional feature spaces,
design practical engines for application of the described
models, integrate the latter with and within deep learning concepts and implementations,
and consider other important activities that are also related to model building (e.g.~planning,
diagnoses, learning, recommendation, etc.).  A particularly fundamental remaining question 
is if real-world entities have some precisely well-defined respective models that are 
completely independent of humans and, if so, how these models could be somehow inferred.  

Going back to the initial problem of how living beings and other modeling agents
including humans and machines may have overcome the so-called information schism,
the concepts and possibilities discussed in Section~\ref{sec:collab} can be understood
to have provide respective insights.  More specifically, the information schism seems to have
been circumvented by creative modeling frameworks which may well be directly related to those
described in the present work, especially in the sense of providing means for progressive
adaptation to the environment demands, and by establishing communications and
collaborations between the  involved modeling agents.
These insights may also extend to smaller scales, as at the molecular 
or cellular level, such as regarding the onset of multicellular
organisms which could be understood as a interconnected body of individual specialized
agents exchanging mass, energy and information.

All in all, it is felt that the developed modeling approach has potential for satisfying 
at least partially many of the requisites listed in Section~\ref{sec:requisites}.
 At the same time, it is important to keep in mind that several of the suggested concepts
and models are still subject of further formalization, validation, and extensions.
Perhaps one of the main results of the described developments ultimately resides in the
fact that while it is interesting to find out if a given dataset belongs to a specific category,  it
may be even more interesting to have a complete objective description, in the form of
a respective model, providing possible explanations of why this takes place.

\begin{figure*}[]  
\begin{center}
   \includegraphics[width=0.7\linewidth]{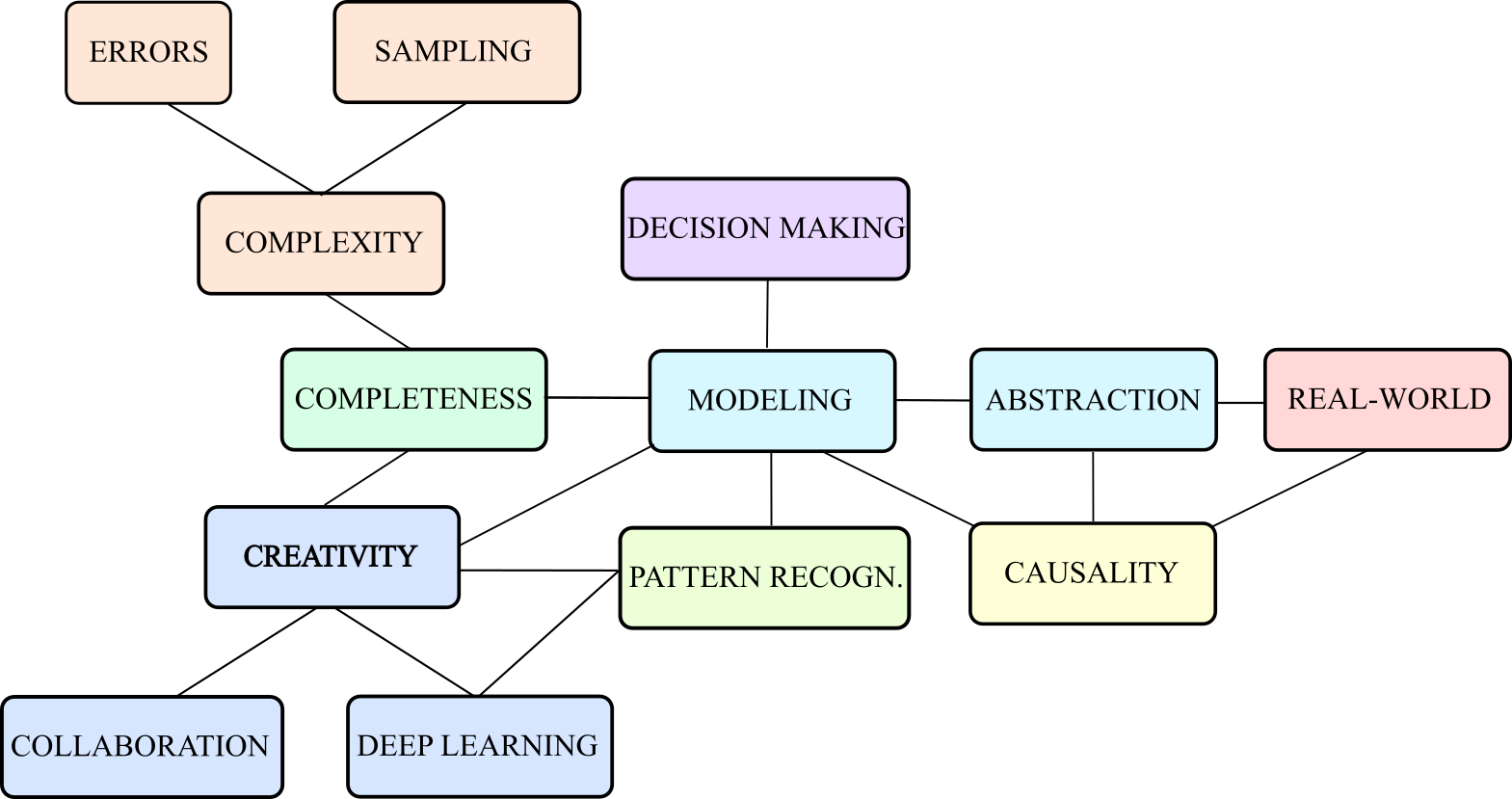}  
    \caption{Graph abstract interrelating the main subjects of this work.  The main concepts and areas covered here, 
    as well as some of their most important interconnections.  
    The main issues associated the modeling challenges are shown in orange, while some
    possible concepts that can contribute for solving those issues are represented in light blue.  }
    \label{fig:diagram}
    \end{center}
\end{figure*}

We conclude by presenting in Figure~\ref{fig:diagram} a graph abstract illustrating
the main concepts addressed in the present work, as well as some of their most
relevant interconnections.

\vspace{3cm}

\vspace{0.7cm}
\emph{Acknowledgments.}

Luciano da F. Costa
thanks CNPq (grant no.~307085/2018-0) and FAPESP (grant 15/22308-2).  
\vspace{1cm}

\bibliography{mybib}
\bibliographystyle{unsrt}

\end{document}